\documentclass[9pt,twocolumn,twoside]{osajnl}
\usepackage{times}
\usepackage{amssymb}
\usepackage{multirow}
\usepackage{mathtools}
\usepackage{subfloat}
\usepackage{bbm}
\usepackage{dsfont}
\usepackage{algorithm}
\usepackage{algpseudocode}
\usepackage{pbox}
\usepackage{pifont}
\newcommand{\xmark}{\ding{55}}
\usepackage{placeins}
\journal{ol} 
\setboolean{shortarticle}{false}
\usepackage{tabularx}
\usepackage{booktabs}
\usepackage{threeparttable}
\usepackage{times}
\usepackage{graphicx}
\usepackage{subcaption}
\usepackage{units}
\usepackage{url}

\usepackage{amsmath}

\usepackage[printonlyused]{acronym}

\acrodef{RL}{reinforcement learning}
\acrodef{TO}{trajectory optimization}
\acrodef{PPO}{Proximal Policy Optimization}
\acrodef{MLP}{multilayer perceptron}
\acrodef{MPC}{model predictive control}
\acrodef{SOTA}{state-of-the-art}
\acrodef{TAMOLS}{terrain-aware motion generation for legged systems}
\acrodef{DTC}{Deep Tracking Control}
\acrodef{IK}{inverse kinematics}
\acrodef{WBC}{whole-body control}

\ifx\isReview\undefined
    \newcommand\highlight[1]{#1}
\else
    \usepackage{xcolor}
    \newcommand\highlight[1]{\textcolor{red}{#1}}
\fi

\title{Humanoid Mobility via Attention-based Contacts}

\author{
Junzhe~He$^{1*}$, 
Kehan~Wen$^{2*}$,
Chong~Zhang$^{1,4*}$, 
Clemens~Schwarke$^{1,5}$,
Zewei~Zhang$^{1}$,
Kaixian~Qu$^{1}$,
Siyuan~Hu$^{3}$,
Matthias~Müller$^{3}$,
Vladlen Koltun$^{3}$,
Marco~Hutter$^{1}$ 
\\
\normalsize{$^{1}$Robotic Systems Lab, ETH Zurich, 8092 Zurich, Switzerland.} \\
\normalsize{$^{2}$ Previously at $^{1}$, now at Department of Electrical and Computer Engineering, National University of Singapore, Singapore.} \\
\normalsize{$^{3}$Apple \qquad $^{4}$ETH AI Center \qquad $^{5}$NVIDIA} \\
\normalsize{$^{*}$Equal contribution.}
}

\begin{abstract}
Humanoid mobility over complex terrain requires accurate foot placement and coordinated whole-body motion capable of robust adaptation. Reinforcement learning (RL) policies trained without motion references generalize well across terrains, yet eliciting effective whole-body coordination through reward design and exploration alone remains challenging. In contrast, reference-based methods yield natural motions but struggle to adapt beyond the predefined trajectories in their training data. We introduce a learning framework that bridges these two approaches by using motion data strictly as a training prior. An explicit, self-supervised, attention-based contact planner predicts future body–environment contacts, while a contact-conditioned tracker translates these plans into coordinated whole-body motion. Together, they establish contact as an interface between sensory observations and control actions. This abstraction allows the policy to internalize the contact-aware structure of locomotion rather than directly replicating kinematics. Consequently, a single policy can exploit data-driven motion priors for natural whole-body coordination while retaining the flexibility to discover environment-dependent behaviors through reinforcement learning. Building on the learned contact planner and tracker, a subsequent fine-tuning stage bootstraps their acquired capabilities to explore environment variations beyond those covered by the original motion data. Operating entirely from onboard observations without any reference inputs at deployment, this framework enables versatile humanoid mobility in challenging environments.

\end{abstract}

\begin{document} 
\maketitle 

%
\section*{Introduction}
\label{sec:intro}
Deploying humanoid robots in human-centric environments requires capabilities well beyond stable bipedal walking. Traversing cluttered and geometrically constrained environments often demands stepping across sparse footholds, climbing onto obstacles, crawling through confined spaces, or leveraging the upper body for support. These multi-contact behaviors fundamentally diverge from conventional locomotion. Instead of adhering to a prescribed pattern of foot contacts, the robot must continuously determine the spatial, temporal, and anatomical distribution of its environmental interactions. Achieving this capability while remaining robust to disturbances and generalizing to previously unseen terrain remains a central challenge for humanoid mobility.

Reinforcement learning (RL) has proven highly effective for teaching quadrupedal and bipedal robots to swiftly navigate rough~\cite{Lee2020, Zhibin20, kumar2021rma, Rudin2021, Miki2022, Agarwal2022, Suyoung23}, sparse~\cite{Tsounis2020, jenelten2023dtc, ame, start}, and even parkour~\cite{zhuang2023robot, anymal_parkour, cheng2024parkour, luo2024pie, parkourinthewild, Hyeongjun25, JunGill26, ame2} terrains without reference motions. Similar principles have increasingly been applied to humanoids, enabling robust behaviors that adapt well to terrain variations~\cite{Radosavovic24, gu2024humanoid, zhuang2024humanoid, zhang2024wococo, long2024internalmodel, wang2025beamdojo, xie2025humanoid, sun2025humanoid, ame, ame2, zhu2026hiking, zhang2026rpl, TAGA, fu2026global}. However, extending this paradigm to whole-body motion control introduces a massive combinatorial space. The robot must simultaneously determine which anatomical parts establish contact, the precise spatial and temporal distribution of those contacts, and the coordinated kinematics of the remaining degrees of freedom. Encoding these complex interactions indirectly through task objectives requires extensive reward shaping, where minor adjustments can yield drastically different contact strategies~\cite{wang2026apex}. Consequently, while pure RL provides a powerful mechanism for robustness, reward signals alone offer insufficient structure for efficiently discovering complex whole-body behaviors.

\begin{figure*}
\centering
    \includegraphics[width = 7.3in]{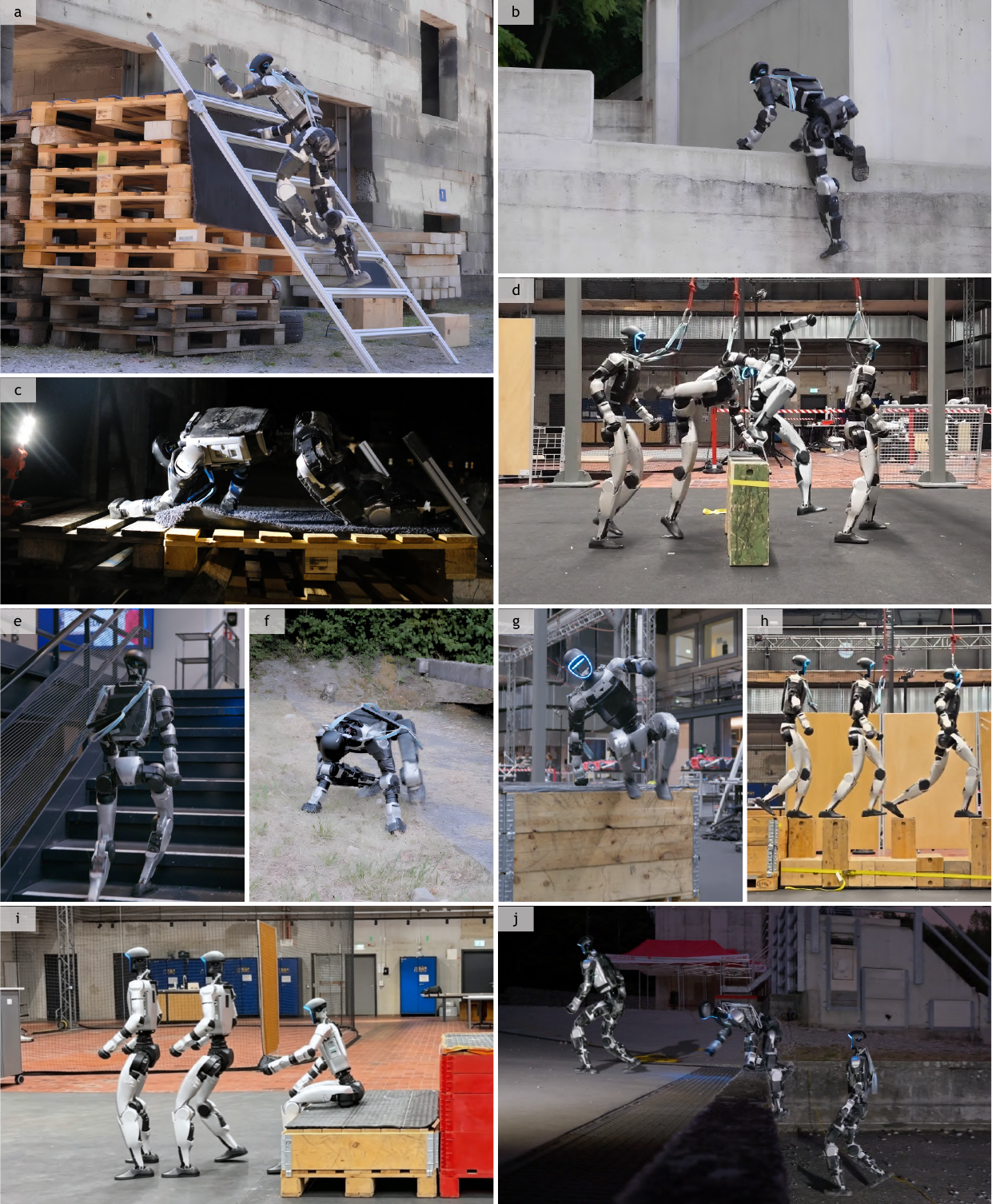}
    \caption{\textbf{\highlight{Agile and robust humanoid mobility in diverse environments.}} Our holistic controller enables a humanoid robot (the Unitree G1) to dynamically traverse diverse and challenging environments, including (a) ladder climbing, (b,j) stage climbing, (c) crawling, (d) vaulting over thin walls, (e) stair climbing, (f) standing up from the ground, (g) jumping down, (h) walking over stepping stones, and (i) sitting down, using a single generalist policy.}

    \label{fig:fig1}
\end{figure*}

An alternative is to provide this structure directly through reference motions. Motion-imitation methods such as DeepMimic~\cite{Peng2018} reframe behavior discovery as a constrained tracking problem, enabling highly dynamic and coordinated whole-body motions. More recent systems extend this idea to humanoids using large motion datasets, retargeting, and learned motion priors~\cite{fu2024humanplus, he2024learning, xie2025kungfubot, ze2025twist, li2025clone, videomimic, liao2025bydmmc, yang2026omni, ZEST, SONIC}. Such approaches naturally produce coordinated and human-like behaviors because the reference specifies not only the desired task outcome but also how the entire body should achieve it. Yet a tracking controller can only execute a suitable motion if an appropriate reference can first be generated for the current environment. During deployment in previously unseen environments, this shifts the central problem from control to reference generation.

One solution is therefore to generate these references online. Model-based approaches can use trajectory planners to provide dynamically feasible motions that are subsequently tracked by a learned controller~\cite{jenelten2023dtc, HoJaelee2024}, given a user-specified velocity or goal command. This decomposition combines explicit reasoning about environmental geometry with the robustness of learned control. Yet it also makes deployment dependent on a whole-body planner, which can be computationally expensive and inherits the modeling assumptions and failure modes of the planning system. For highly dynamic locomotion with intermittent multi-contact interactions, generating accurate full-body trajectories in real time remains particularly challenging.

Recent work has consequently begun replacing model-based reference generation with learned motion generators or planners~\cite{PARC, zewei26, VLK}. These methods provide a promising route toward more expressive whole-body behaviors by learning the distribution of feasible motions from data. Nevertheless, generating complete motion trajectories remains a high-dimensional prediction problem. Generalization to environment configurations not represented in the motion data often requires substantial data augmentation, and small trajectory errors can become consequential when steppable or contactable regions are sparse. In such environments, producing a motion that is kinematically plausible is insufficient: the predicted interaction with the environment must also be physically accurate.

A related line of work attempts to close the gap between reference-based training and autonomous deployment by composing, augmenting, or distilling reference motions into perceptive policies~\cite{php, xloco, HIL, HIL_hardware, light_loco}. Wu et al.~\cite{php}, for example, use motion matching to compose retargeted human skills into obstacle-adaptive reference trajectories, train tracking experts, and subsequently distill these behaviors into a vision-based policy. Wang et al.~\cite{xloco} introduce a generalist humanoid locomotion framework that trains multiple specialist policies for diverse skills and distills their expertise into a single vision-based policy through case-adaptive specialist selection, enabling unified locomotion control without reference motions at deployment. Wang et al.~\cite{HIL, HIL_hardware} propose a hybrid imitation learning framework that combines reference motion tracking with adversarial imitation learning, enabling a unified controller to preserve the naturalness of motion data while improving skill composition and adaptation to procedurally generated environments. Chen et al.~\cite{light_loco} introduce a single perceptive whole-body policy that autonomously transitions between conventional locomotion and parkour skills without reference motions or skill labels at deployment, using terrain-conditioned reference augmentation during training to acquire diverse whole-body behaviors from a small set of motion seeds. These approaches demonstrate that motion data can provide valuable priors for highly dynamic humanoid behaviors. However, they require machinery for adapting and composing full-body motions across environmental variations. As the diversity and required geometric precision of the terrain increase, constructing references that cover the relevant combinations of obstacle geometry, contact location, and skill transition becomes increasingly difficult. This raises a fundamental question: how much of a reference motion does a robot actually need in order to learn generalizable whole-body locomotion? 

We argue that the essential information lies not in the complete kinematic-space trajectory, but in its interaction with the environment. Consider how humans traverse irregular natural terrains. Human locomotion does not rely solely on reacting to the terrain immediately beneath the feet; instead, visual information is gathered prospectively to determine suitable future interactions with the environment. Matthis et al.~\cite{gaze} studied gaze and full-body kinematics during walking over natural outdoor terrain and showed that humans adapt their gaze behavior according to terrain complexity while maintaining a consistent temporal look-ahead strategy. Their findings reveal a tight coupling between perception of the upcoming terrain and the subsequent decisions on contact positions. These observations suggest that successful locomotion over geometrically constrained terrain depends critically on identifying where future contacts should occur, rather than prescribing the complete body trajectory leading to them.

The importance of contact has also been recognized in recent learning-based control methods. Zhang et al.~\cite{zhang2024wococo} condition whole-body humanoid control on sequential contact objectives, demonstrating that contact specifications can effectively coordinate diverse whole-body behaviors. Omar et al.~\cite{learn_to_act} use contact-centric representations to facilitate multi-task robot learning. More recently, Chen et al.~\cite{SceneBot} incorporate scene-interaction contacts as prompts for general humanoid whole-body motion tracking. Yu et al.~\cite{OmniContact} introduce contact flow, combining key body trajectories with temporal contact signals, as a shared interface for executing and composing long-horizon humanoid loco-manipulation skills. Together, these works demonstrate that contacts provide a compact and effective abstraction for specifying complex robot behaviors. 

This perspective becomes even more relevant for whole-body mobility. When traversing sparse and complex terrain, the critical decisions extend beyond selecting footholds: the robot must determine which parts of the body should interact with the environment and where these interactions should occur. Once these contacts are specified, however, there may exist many valid whole-body trajectories that realize them. A hand contact on the top of a ledge, for example, constrains an essential interaction with the environment without uniquely determining the trajectories of the arms, torso, and legs. We therefore propose contact as an intermediate representation between terrain perception and whole-body control. Compared with a full-body reference trajectory, contacts retain the geometrically critical information required for terrain interaction while leaving the policy freedom to adapt the motion connecting them.

This representation also changes the role of motion data. Rather than using demonstrations as trajectories that the robot must reproduce, we use them as training-time guidance for learning which body–environment interactions are useful. Reference motions provide examples of effective contacts during training, while reinforcement learning determines how those contacts should be realized under different robot states and terrain configurations. The resulting policy is therefore not required to remain close to a reference trajectory. More importantly, the reference is removed entirely at deployment: given only the robot state, terrain observation, and a user-specified goal, the policy must autonomously determine appropriate contacts and execute the corresponding whole-body motion. In this way, demonstrations provide structure to the otherwise difficult exploration problem of whole-body locomotion without constraining the policy to the distribution of full-body trajectories contained in the dataset.

\begin{figure*}
\centering
\includegraphics[width=7in]{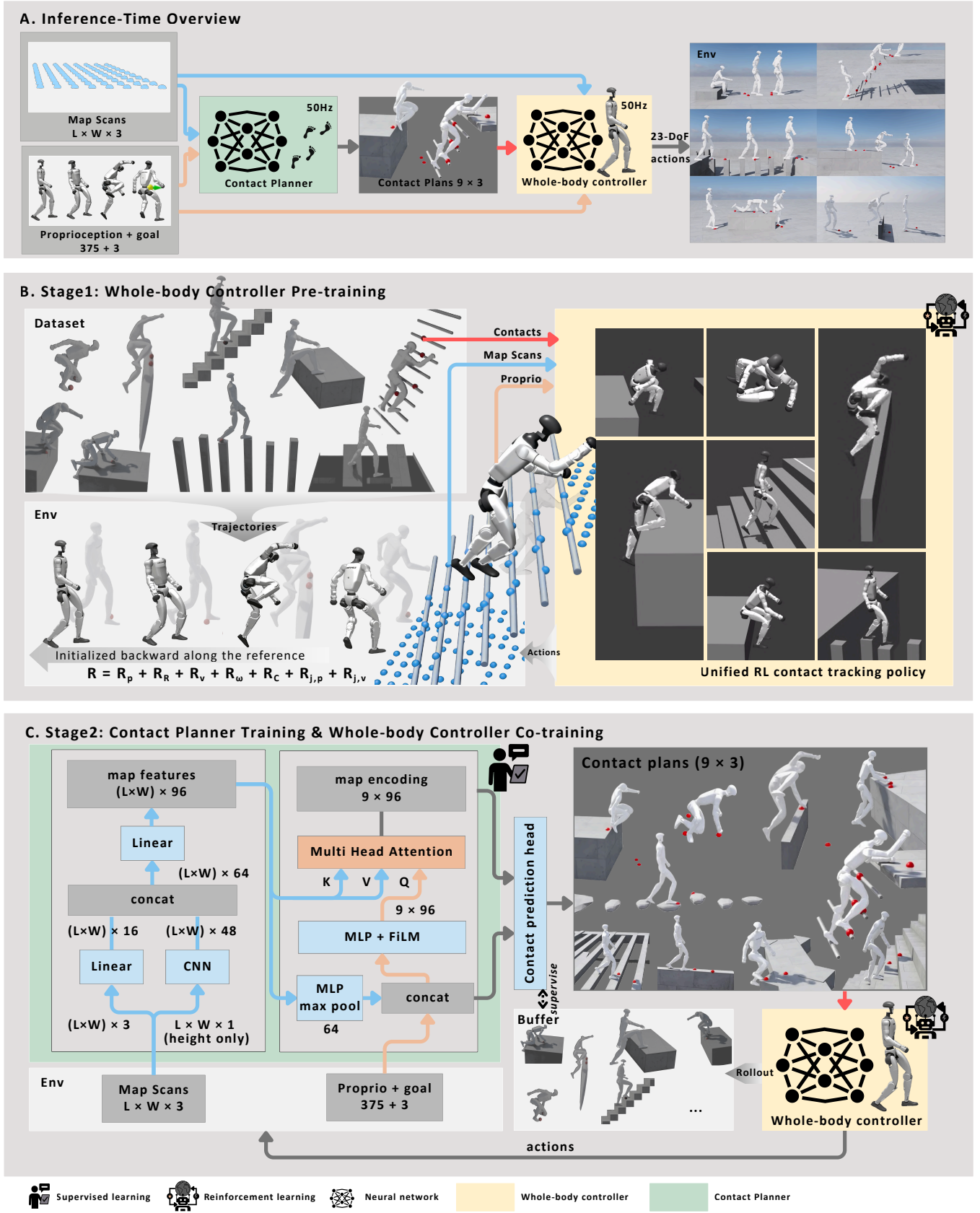}
\caption{\textbf{\highlight{Overview, part I.}} \textbf{(A)} Inference-time pipeline. \textbf{(B)} Stage 1: whole-body controller pretraining. Reference trajectories are sampled from the dataset. Training follows a backward curriculum: the robot is initialized at progressively earlier states along the reference trajectory, advancing to an earlier state only after the current one has been successfully learned. $R_{P}, R_{R}, R_{v}, R_{\omega}$, $R_{C}$,  $R_{j,p}$, and $R_{j,v}$ denote the position, rotation, linear velocity, angular velocity, contact, joint position, and joint velocity tracking reward, respectively. \textbf{(C)} Stage 2: attention-based contact planner. The planner predicts contact plans by attending to map observations, conditioned on robot proprioception and the goal command. It is trained in a supervised manner using ground-truth contact labels from a motion buffer containing rollout trajectories generated by the whole-body controller. The tracker is co-trained using the predicted contacts from the planner.
}
\label{fig:overview}
\end{figure*}

\begin{figure*}
\centering
\includegraphics[width=7in]{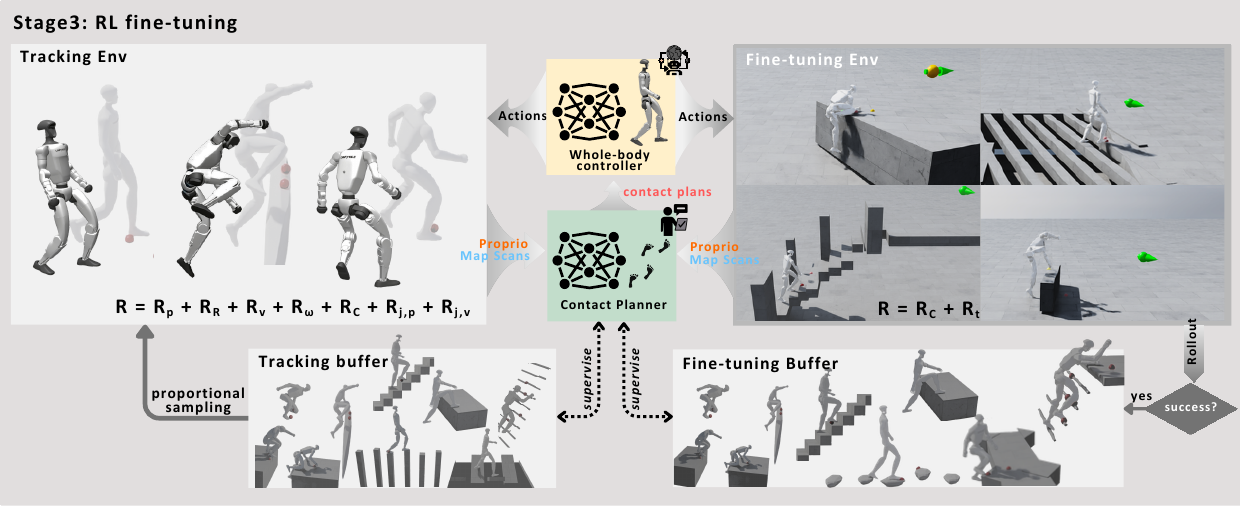}
\caption{\textbf{\highlight{Overview, part II.}} Stage 3: RL fine-tuning. Training combines reference tracking on motions sampled from the demonstration dataset (left) with goal-reaching rollouts on procedurally generated terrains outside the demonstration distribution (right). In the goal-reaching environment, the contact planner predicts contact plans from map scans, robot proprioception, and the goal command; the whole-body controller executes these plans and is fine-tuned with RL using task ($R_t$) and contact-tracking ($R_C$) rewards. Successful rollouts are stored in the fine-tuning buffer and subsequently used as additional supervision for the contact planner, forming a bootstrapping loop in which planner-generated contacts are executed, validated through successful interaction, and recycled to improve the planner. Reference trajectories are sampled proportionally from the original dataset during fine-tuning to retain the previously learned tracking behaviors.} 
\label{fig:stage3}
\end{figure*}

To autonomously infer these contacts from terrain observations, we build upon attention-based map encoding (AME)~\cite{ame, ame2}. AME uses robot proprioception to condition attention over a local terrain representation, allowing the policy to selectively extract terrain features relevant to its current locomotion state. In previous work, this mechanism was learned implicitly and was shown to focus on steppable terrain regions associated with future footholds. Here, we exploit an alternative capability of this architecture: attention provides a natural interface for explicit contact planning. Rather than using attention only to encode terrain into a latent representation for control, we use it to identify task-relevant contact locations on the terrain map. Specifically, our contact planner attends to terrain points conditioned on the robot's proprioception and a user-specified goal and predicts explicit contact locations directly on the map. This formulation tightly couples when and where the robot should interact with the environment to its evolving physical state. The planner can therefore shift its attention as the robot moves, selecting different terrain regions for feet, hands, and other body contacts according to both local geometry and the current proprioceptive state. This design is particularly suited to sparse terrain, where a small error in a predicted contact can distinguish a feasible interaction from a failure. At the same time, predicting contacts rather than complete whole-body trajectories substantially reduces the dimensionality of the planning problem.

Based on these insights, we introduce a framework that uses an explicit, self-supervised attention-based contact representation as the interface between perception and whole-body control.  During training, reference motion data guides the discovery of effective body-environment interactions. Rather than tracking full reference trajectories, the system abstracts the motion data into a task-relevant contact structure. The policy then leverages RL to physically realize these target contacts while optimizing for broader locomotion objectives. This formulation preserves the primary advantage of motion data by providing structured guidance for coordinated whole-body behavior without restricting the agent to strict kinematic reproduction.

This contact-centric representation has several important properties. First, it reduces the dimensionality of the guidance compared to full-body trajectory tracking. Because multiple dynamically feasible motions can realize the same contact objective, the contact representation constrains only the task-relevant interactions while leaving the policy free to explore and optimize the whole-body motion between contacts. Second, contact locations are expressed directly with respect to the environment, providing the geometric precision needed when feasible support regions are sparse. Third, training remains simple: instead of constructing rewards that separately induce climbing, crawling, stepping, and other behaviors, reference contacts can be incorporated through a common contact-tracking objective. Finally, because reference trajectories are used only as training-time guidance, deployment requires neither motion retrieval nor a model- or learning-based motion generator. Terrain diversity can therefore be explored directly through RL rather than covered through exhaustive manual augmentation of reference trajectories.

To obtain diverse and physically feasible whole-body references for training, we use a motion-retargeting pipeline~\cite{zewei26}. It transfers human motion data to the target humanoid while accounting for kinematic constraints. The resulting trajectories, together with their body–environment contact labels and positions, provide contact-aware references as training-time guidance for learning whole-body locomotion.

In summary, this paper introduces a three-stage training framework that progressively transfers knowledge from motion data to a whole-body control policy, with the following contributions: \textbf{1)} A contact-informed whole-body tracking training pipeline where the policy is conditioned on contact plans instead of whole-body trajectories. \textbf{2)} A supervised learning formulation for contact planning, leveraging human motion data to capture rich and diverse contact strategies, and an attention-based contact planner (ACP) that predicts contact plans for nine body parts (hands, elbows, knees, feet, and pelvis) on rough and discontinuous terrain given a user-specified goal. \textbf{3)} An RL fine-tuning stage bootstrapping on planner-generated contact plans that bridges data-driven imitation and reward-driven exploration, enabling generalization to unseen terrains. \textbf{4)} A LiDAR-based neural mapping framework that accurately captures fine-grained geometry, including small and thin structures such as ladder rungs, for precise whole-body locomotion.

Through three training stages -- contact tracking, contact planning, and RL fine-tuning -- our framework progressively reduces its reliance on motion data while preserving the coordinated whole-body behaviors acquired from demonstrations. By isolating task-relevant contact structure, this framework yields a unified generalist policy capable of versatile, coordinated whole-body motion control in previously unseen environments using only onboard sensing (Figure \ref{fig:fig1}). Figures \ref{fig:overview} and \ref{fig:stage3} provide a visual overview of the method.

\section*{Results}
\subsection{Agile and generalized whole-body locomotion}
\label{sec:wbloco}
\begin{figure*}
\centering
\includegraphics[width=7.3in]{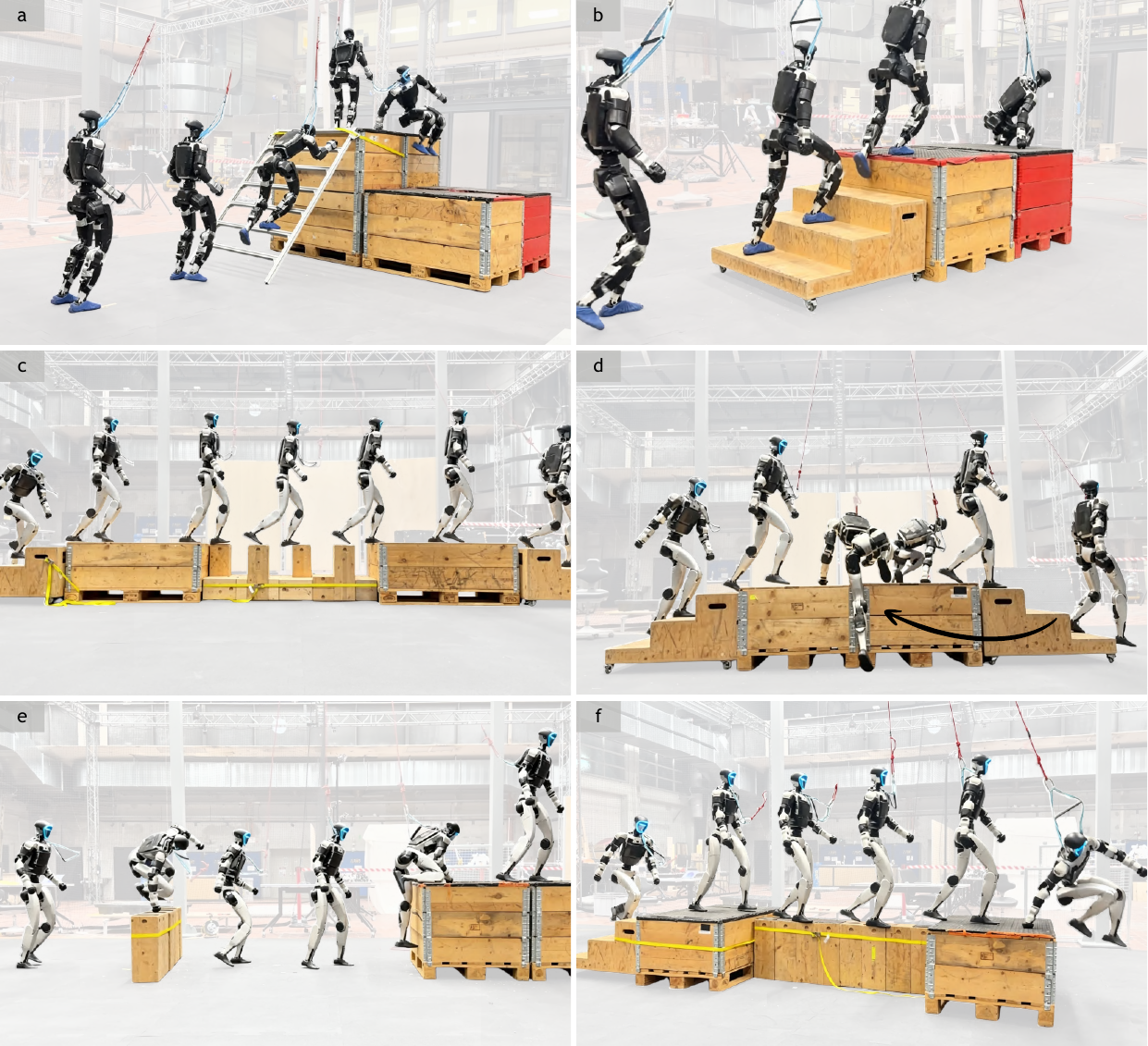}
\caption{\textbf{\highlight{Compositional whole-body motion control.}} \textbf{(a)} Walking - ladder climbing - jumping down. \textbf{(b)} Walking - stairs climbing - jumping down. \textbf{(c)} Stairs climbing up - stepping stones - stairs climbing down. \textbf{(d)} Stairs climbing up - stairs climbing down - box climbing up - jumping down.  \textbf{(e)} Walking - vaulting - box climbing up. \textbf{(f)} Stairs climbing up - beam - jumping down. All motions deployed with a single generalist policy.}
\label{fig:results_real}
\end{figure*}

\begin{figure*}
\centering
\includegraphics[width=7.3in]{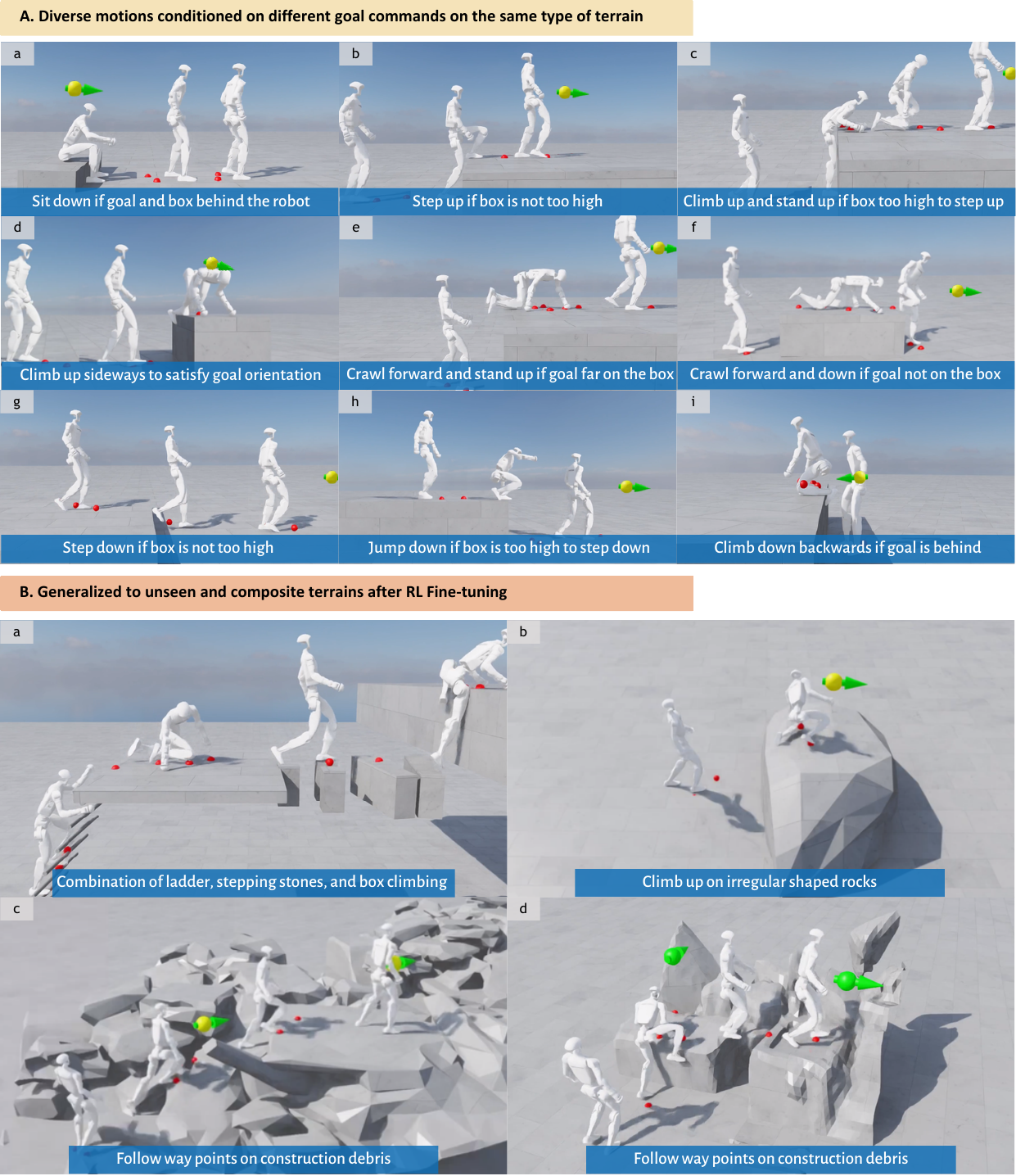}
\caption{\textbf{\highlight{Simulation results.}} The red dots indicate contact plans; the sphere and arrow indicate goal position and orientation. \textbf{(A)} The robot selects different interaction strategies on the same type of terrain depending on the desired goal position and orientation. \textbf{(B)} The fine-tuned policy generalizes to composite terrains and unseen terrains.}
\label{fig:generalization}
\end{figure*}

We tested our unified generalist policy on the Unitree G1 humanoid robot through extensive real-world experiments (Figures~\ref{fig:fig1} and~\ref{fig:results_real}) across a diverse range of challenging indoor and outdoor terrains. Operating without reference motions at deployment, the controller achieves precise, coordinated whole-body locomotion by autonomously adapting its contact strategy to the perceived terrain. The resulting behaviors include stair, ladder, and box climbing, sitting down, standing up from the ground, sparse and composite terrain traversal, and other whole-body interactions. 
Importantly, successful execution requires more than recognizing the surrounding geometry: the policy must also select an appropriate interaction strategy according to user-specified goal. Consequently, the same environment can elicit substantially different whole-body behaviors depending on the desired goal position and orientation as well as the geometry. For example, as shown in Figure \ref{fig:generalization}A, when approaching a box, a forward-facing goal behind the robot can let the robot sit down (a), whereas a forward-facing goal in front of the robot on top of a low box prompts the robot to step onto it (b). If the box is too high, the policy instead leverages a whole-body climbing strategy that uses the hands for additional support (c). When the desired orientation points to the left or right, the policy reorients its body and performs a sideways climbing maneuver (d). When the goal is located farther across the top surface, the robot climbs onto the box, crawls forward toward the target, and then stands up (e). If the goal is on the other side of the box, the robot climbs up, crawls forward, then steps down (f). Similarly, different motions can be generated for getting down from a box given different heights and goal commands (g-i). These examples demonstrate that our unified generalist policy does not simply associate a terrain category with a fixed motion primitive, but jointly interprets environment geometry and goal commands to synthesize appropriate contact patterns and whole-body motions online.

The unified policy further generalizes to unseen and composite terrains, where multiple terrain types and interaction modes are encountered sequentially, as shown in Figure \ref{fig:generalization}B. In these scenarios, the robot must not only execute individual skills but also autonomously determine how to connect them as the terrain and goal evolve. The policy synthesizes appropriate motion transitions between behaviors without manually designed transition motions, explicit skill-switching mechanisms, or augmentation of the motion dataset for specific terrain compositions. For example, it can transition continuously from walking to climbing, crawling, standing, or other contact-rich behaviors as required by the local geometry and the user-specified goal. In Figures \ref{fig:generalization}B-a and \ref{fig:generalization}B-b, only one goal is specified. For the unstructured terrains shown in Figures \ref{fig:generalization}B-c and \ref{fig:generalization}B-d, these goals are provided as manually specified intermediate waypoints to guide local traversal. (Global navigation is beyond the scope of this work.) The resulting policy successfully traverses diverse unseen terrains while maintaining its original ability without requiring additional reference motions or terrain-specific motion augmentation. This capability allows a single policy to compose previously learned interaction strategies into continuous whole-body locomotion over terrain configurations not represented in the original motion dataset. Together, these results demonstrate precise, robust, and goal-conditioned whole-body humanoid locomotion that generalizes beyond both the motions and terrain configurations encountered during training. Detailed simulation-based analysis is provided in Section~\ref{sec:evaluation}.

\subsection{Failure Recovery}
\begin{figure*}
\centering
\includegraphics[width=7.3in]{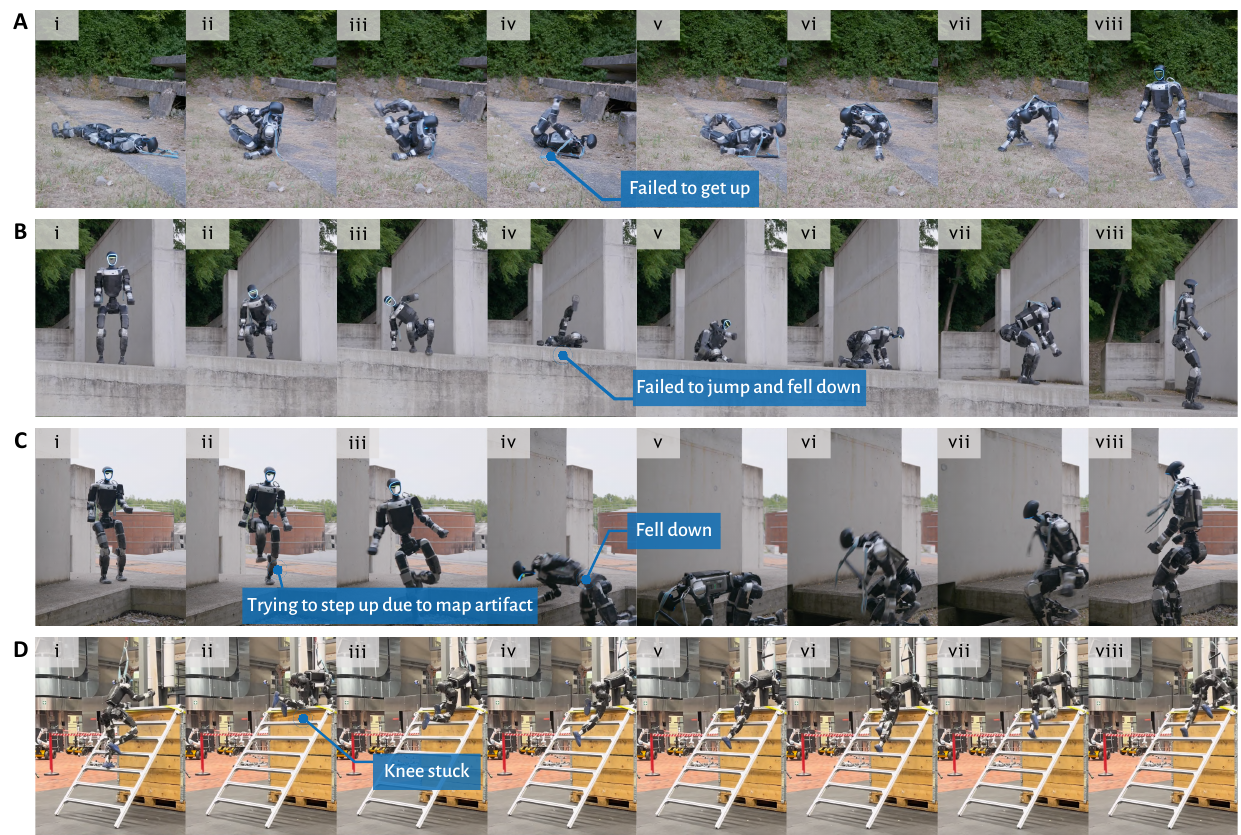}
\caption{\textbf{\highlight{Failure recovery.}} Our generalist controller enabled failure recovery behaviors including but not limited to: \textbf{(A)} re-attempting to stand up if failed due to uneven ground, \textbf{(B)} and \textbf{(C)} standing up after falling down, and \textbf{(D)} recovering from failures (e.g., knee stuck, slipped down from a rung, etc)}
\label{fig:emergent}
\end{figure*}

Our learned policy demonstrates strong robustness and emergent failure-recovery capabilities on the real robot, as shown in Figure~\ref{fig:emergent}. Figure~\ref{fig:emergent}A shows that, after an unsuccessful stand-up attempt on uneven terrain, the policy autonomously makes repeated attempts until it successfully recovers to a standing configuration. Figures~\ref{fig:emergent}B and \ref{fig:emergent}C demonstrate recovery from unexpected falls: when the robot fails to land a jump-down maneuver or is tipped over because of a perception failure, the policy recovers from the ground and resumes locomotion. Figure~\ref{fig:emergent}D further shows recovery during ladder climbing, where the robot can escape from a stuck configuration and repeatedly attempt to climb onto the platform. Notably, these recovery trajectories are not explicitly represented in the original motion dataset, suggesting that the policy can synthesize corrective behaviors beyond the provided demonstrations to recover from failures encountered during deployment.

\subsection{Simulation-based Evaluations}
\label{sec:evaluation}

\begin{figure*}[p]
\centering
\includegraphics[width=\textwidth,height=0.88\textheight,keepaspectratio]{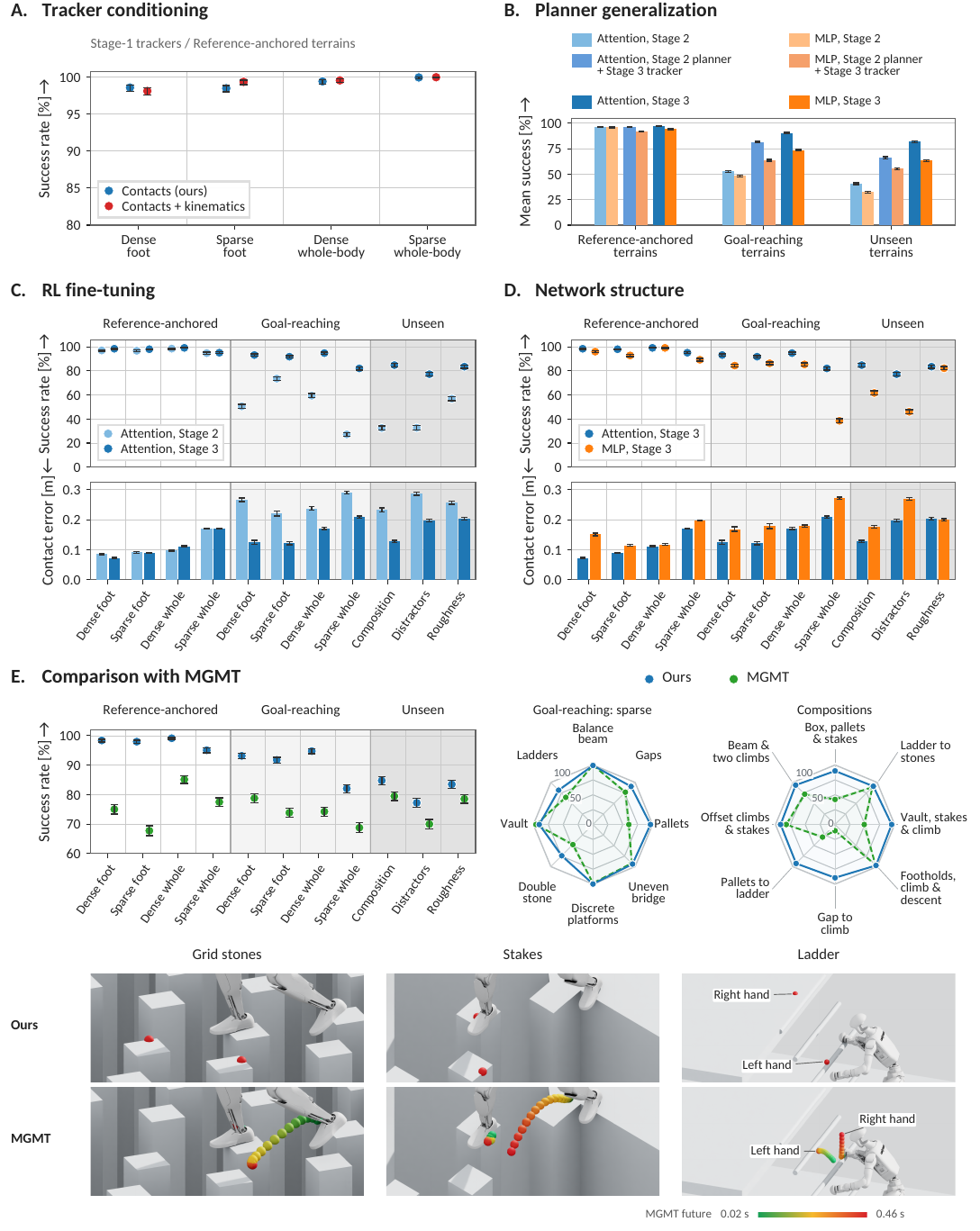}
\caption{\textbf{Simulation-based evaluations.} \textbf{(A)} Stage-1 trackers using contacts alone or contacts plus reference kinematics. \textbf{(B)} Attention and MLP systems under three configurations: Stage~2, a frozen Stage-2 planner paired with a Stage-3 tracker, and the fully fine-tuned Stage-3 system. Bars show mean success, with terrain categories weighted equally. \textbf{(C)} Attention-based systems before and after RL fine-tuning. \textbf{(D)} Attention versus MLP after fine-tuning. In C and D, the upper plots show success rates and the lower plots show contact placement errors; \textit{foot} and \textit{whole} denote foot-only and whole-body support. \textbf{(E)} Comparison with MGMT~\cite{zewei26} across terrain categories, individual sparse terrains, and eight composite courses. Each radar axis uses 1,000 episodes per method. The images compare predictions from the same robot state: red markers indicate our contact targets, while MGMT's predicted foot and hand positions progress from green (earlier) to red (later). Whiskers and shaded bands show 95\% confidence intervals.}
\label{fig:ablation_v14}
\end{figure*}

We use simulation to examine whether contact commands provide sufficient guidance for whole-body tracking, how learned contact plans transfer beyond the reference terrains, and how goal-reaching RL fine-tuning and attention-based planning contribute to this generalization (Figure~\ref{fig:ablation_v14}).

We distinguish three terrain groups by their usage in different training stages. \textit{Reference-anchored terrains} are associated with reference motions used in Stage~1/2, and test retention of demonstrated behaviors. \textit{Goal-reaching terrains} are the procedurally generated terrains used for reference-free fine-tuning in Stage~3 and provide a zero-shot transfer test for Stage-2 policies. \textit{Unseen terrains} introduce obstacle composition, task-irrelevant distractors, and surface roughness to test skill transitions, rejection of irrelevant geometry, and traversal of irregular support surfaces, respectively.

The first two groups each contain four categories, defined by support density and the body parts required for support. Dense foot-only terrains provide continuous support for walking, such as stairs and slopes; sparse foot-only terrains require precise foot placement, as on stepping stones and narrow beams. Dense whole-body terrains provide broad surfaces for additional hand or knee support, as in box climbing and get-up transitions; sparse whole-body terrains offer only discrete contacts for multiple links, as in ladder climbing and vaulting. This division distinguishes the placement precision required by sparse support from the multi-limb coordination required by whole-body interactions. Training and evaluation terrain definitions are provided in the
\hyperref[sec:supp_training_terrains]{supplementary training details}
and \hyperref[sec:supp_evaluation_terrains]{supplementary evaluation details},
respectively. 

Category-level comparisons pool 3,000 episodes per method and category across three evaluation seeds. The terrain-specific comparisons in Figure~\ref{fig:ablation_v14}E use 1,000 episodes per method and radar axis, with the ladder axis pooling four variants. The eight curated composite courses are evaluated separately from the unseen-terrain benchmark. For category-level evaluations, success requires reaching within 0.5\,m of the active goal horizontally; for sitting, standing, and get-up transitions, the criterion is instead a vertical error below 0.1\,m. On the curated courses, success additionally requires reaching the target height and passing through the prescribed course without entering the pit or leaving the traversal lane.
Contact placement error measures the Euclidean distance between commanded and realized contact positions at contact onset, averaged over valid contact events. We report Wilson 95\% confidence intervals for success and episode-cluster-robust 95\% intervals for contact error. These quantify evaluation uncertainty rather than variability across training runs. Episodes with invalid policy outputs count as failures and are excluded from contact error. Sensor noise, physics randomization, and action delay are disabled to isolate terrain-dependent performance.

\subsubsection{Ablation study on tracker design}
On reference-anchored terrains, we compare the Stage-1 contact-conditioned tracker with a variant that additionally receives reference root and joint kinematics. Both receive the same reference contact commands and retain reference-motion guidance during training. Contacts alone achieve at least 98.5\% success in every category, with no consistent improvement from the additional kinematic inputs (Figure~\ref{fig:ablation_v14} A). This demonstrates that contact plans alone suffice to guide the whole-body skills in our dataset, without requiring high-dimensional full-body reference kinematics as policy inputs.

\subsubsection{Ablation study on planner architecture and generalization}
We compare our attention-based planner with a multilayer perceptron (MLP) that predicts per-body contact plans over the flattened height scan. The two variants have comparable parameter counts and share the same contact interface, supervision, and terrain-point decoder; their architectures and parameter counts are reported in Table~\ref{tab:supp_networks}. 

After Stage~3 fine-tuning, attention achieves higher success than the MLP across the goal-reaching terrain categories and shows clear gains on unseen composition and distractors (Figure~\ref{fig:ablation_v14}D). Its largest advantages occur on sparse whole-body terrain and with distractors, where the planner must identify precise supports or distinguish them from irrelevant geometry. The improvement on composition also indicates better transfer of contact selection to new sequences of familiar obstacles.

To assess whether this architectural advantage persists without planner fine-tuning, we pair each frozen Stage-2 planner with its corresponding Stage-3 tracker (Figure~\ref{fig:ablation_v14}B). The attention-based pairing achieves higher mean success in all three terrain groups. Contact plans learned on reference-anchored terrains can guide an adapted tracker on goal-reaching and unseen terrains without further planner training. Because each planner uses its own adapted tracker, this comparison reflects both planner generalization and planner--tracker compatibility.


\subsubsection{Ablation study on RL fine-tuning}
Goal-reaching fine-tuning extends skills learned from motion data to terrains without corresponding reference motions. For the attention-based system, the Stage~3 policy retains high success on reference-anchored terrains while improving success and contact placement accuracy in every goal-reaching and unseen terrain category (Figure~\ref{fig:ablation_v14}C). The largest gain occurs on sparse whole-body terrain, where success increases by 54.8 percentage points, highlighting the benefit of adaptation when coordinated body contacts must fit constrained supports.

The gains also extend beyond the terrain conditions used for fine-tuning. The largest improvement on unseen terrains occurs on composition, indicating better transitions between familiar interactions in new obstacle sequences. Improvements with distractors and roughness show that fine-tuning also helps the system handle irrelevant geometry and irregular support surfaces. Contact guidance thus allows demonstrated skills to be adapted through reference-free, goal-reaching experience, without a matching full-body trajectory for each terrain configuration or transition.

\subsubsection{Quantitative comparison with MGMT}
We compare our Stage-3 system across the three terrain groups in Figure~\ref{fig:ablation_v14}E with MGMT~\cite{zewei26}, the closest prior work, which couples diffusion-based motion generation with a tracker conditioned on full reference kinematics. 
Our method achieves higher success in all eleven terrain categories. Terrain-specific evaluations show particularly large gains on various sparse terrains where traversal requires accurate placement on discrete supports. 
We further evaluate eight curated composite courses combining sparse footholds with whole-body obstacles. The visualization of the composite terrains can be found in Figure \ref{fig:composite} in the supplementary methods section. Our method achieves higher success on all eight courses, averaging 92.5\% compared to 58.8\% for MGMT. These results demonstrate its ability to combine skills learned on individual terrains and transition between different support and coordination requirements within a single traversal.
The qualitative examples compare our predicted contact targets with MGMT's future foot and hand positions from the same robot state. Across three scenarios, our planner selects contact locations on stepping stones and ladder rungs, providing explicit, terrain-aligned targets for whole-body control. MGMT's predicted trajectories include positions in unsupported regions. Predicting whole-body motion does not, by itself, precisely specify which future limb positions should establish support contacts or ensure their alignment with the terrain.

\subsubsection{Qualitative benchmark with other whole-body control methods}
We qualitatively compare our method with other whole-body humanoid control approaches, including perceptive humanoid parkour (PHP)~\cite{php}, and light-loco-parkour (LLP)~\cite{light_loco}, as shown in Table \ref{tab:qualitative_comparison}. A direct quantitative comparison is challenging because these approaches differ substantially in training data, robot platforms, terrain definitions, perception modalities, and training pipelines. Reproducing each method under identical settings and with our training data is therefore impractical within a reasonable computational and engineering budget. Instead, we report the capabilities demonstrated in the corresponding papers and supplementary materials and compare them qualitatively with the behaviors demonstrated by our method. For behaviors or terrain types that are not demonstrated in the original work, we report the corresponding entry as N/A. This label indicates only that the capability was not demonstrated or evaluated in the available results; it should not be interpreted as evidence that the method is incapable of performing the corresponding behavior. LLP is marked as limited in terrain composition because the paper does not demonstrate compositions of multiple terrain types without clearly defined transition regions like climbing a box from pallets, as shown in Figure \ref{fig:composite}A, climbing a box over a gap (Figure \ref{fig:composite}E), or climbing a ladder directly from pallets (Figure \ref{fig:composite}F), etc. While this does not imply that the method cannot handle such scenarios, its ability to autonomously transition between behaviors on arbitrary terrain compositions remains unclear from the reported results.

\begin{table}[t]
\centering
\begin{threeparttable}
\caption{\textbf{Qualitative comparison of whole-body locomotion methods.}
$\checkmark$: demonstrated; N/A: not demonstrated or evaluated.}
\label{tab:qualitative_comparison}

\small
\setlength{\tabcolsep}{3pt}

\begin{tabularx}{\columnwidth}{@{}Xccc@{}}
\toprule
\textbf{Capability}
& \textbf{PHP~\cite{php}}
& \textbf{LLP~\cite{light_loco}}
& \textbf{Ours} \\
\midrule
Sparse footholds
    & N/A & $\checkmark$ & $\checkmark$ \\
Dense footholds
    & N/A & $\checkmark$ & $\checkmark$ \\
Sparse whole-body
    & N/A & N/A & $\checkmark$ \\
Dense whole-body
    & $\checkmark$ & $\checkmark$ & $\checkmark$ \\
Terrain composition
    & N/A & limited & $\checkmark$ \\
\midrule
No manual data augmentation
    & \xmark & \xmark & $\checkmark$ \\
No manual transition design
    & \xmark & \xmark & $\checkmark$ \\
\midrule
Data length
    & 9.4 min\tnote{$\dagger$} & N/A & 13.2 min \\
\bottomrule
\end{tabularx}

\begin{tablenotes}[flushleft]
\footnotesize
\item[$\dagger$] Before mirroring and motion-matching augmentation;
total duration after augmentation is not reported in~\cite{php}.
\end{tablenotes}

\end{threeparttable}
\end{table}


\section*{Discussion}
Our results suggest that contact can serve as a general interface for humanoid mobility, bridging motion data, sensory observation, and whole-body control. Unlike full-body trajectories, contacts only specify where, when, and with which body part the robot should interact with the environment without prescribing how the entire motion should be executed. This abstraction allows demonstrations to provide structured guidance during training while leaving reinforcement learning the flexibility to explore behaviors on new terrain. As a result, a single policy can compose and transition between diverse whole-body behaviors without reference trajectories, skill labels, or manually designed transitions at deployment.

The attention-based contact planner plays an important role in terms of the generalization capability of our framework. Building on attention-based map encoding~\cite{ame, ame2}, the planner selectively attends to terrain regions relevant to future contacts conditioned on proprioception and goal commands. Rather than coupling a particular terrain configuration to a specific motion, the policy learns to reason about the local geometry that supports each interaction. This allows contact strategies acquired from demonstrated motions to adapt to new terrain geometries and to be recombined on unseen terrain compositions. Goal conditioning further enables different contact strategies to emerge on the same terrain depending on the desired position and orientation. Together, selective terrain attention and explicit contact prediction provide a structured representation that helps transfer whole-body behaviors beyond the terrain in the reference data encountered during training.

Several limitations remain. First, our current perception and contact representations primarily target terrain-centric 2.5-D interactions. Although the LiDAR-based mapping pipeline captures fine geometric structures such as ladder rungs, more general interactions including overhead contacts, hanging, and manipulation will require richer 3D perception and contact representations. Second, the current system also addresses only local goal-conditioned locomotion rather than global navigation. Additionally, the current pipeline cannot handle aerial motions that do not have distinguishable contact sequences. For example, the contact sequences of a backflip and a vertical jump are essentially the same, despite their substantially different whole-body motions, and therefore cannot be distinguished by our current contact representation.

Future work could extend the contact interface to vision-based and general 3D scene representations, allowing contacts with arbitrary surrounding structures rather than predominantly terrain surfaces. Combining such representations with semantic or global planners could enable humanoids to autonomously select goals and interaction strategies over larger environments. Semantic task inputs could also complement the contact representation in cases where different behaviors share indistinguishable contact sequences, providing additional information to specify the intended motion. Ultimately, walking, climbing, crawling, recovery, and manipulation can all be viewed as different sequences of body-environment interactions. Contact therefore offers a common physical abstraction through which humanoids may compose known interaction strategies into new behaviors in previously unseen environments.






\section*{Materials and Methods}

\subsection*{\highlight{Motivation}}
Developing generalizable whole-body motion control exposes a fundamental tradeoff. Pure RL excels at robust, foot-dominant locomotion, but scaling it to whole-body interactions requires complex reward engineering. Conversely, reference-tracking methods provide strong kinematic priors but bind the policy to the source data, necessitating online trajectory generation during deployment. This dependency creates a bottleneck on sparse or unstructured terrains, where precise environmental interactions must rapidly adapt to unseen geometric constraints.

Bridging these approaches does not require transferring complete reference trajectories. Inspired by biological studies showing that human visual attention prospectively targets future foot placements~\cite{gaze}, we propose that the essential information lies strictly in the spatial, temporal, and anatomical distribution of environmental interactions. Contact thus serves as a compact intermediate representation. It retains the geometrically critical structure of demonstrated motions while leaving the high-dimensional kinematics underconstrained, allowing a single contact sequence to specify highly complex behaviors.

Motivated by this, we use future whole-body contacts as the interface between perception, planning, and control (Figure \ref{fig:overview}A). During training, motion data exposes the robot to coordinated interaction strategies, which an attention-based contact planner learns to predict directly from terrain geometry, robot proprioception, and goal commands. Subsequent RL fine-tuning randomizes the terrain without requiring corresponding reference augmentation. This formulation strictly  separates what interactions should occur from how the robot should realize them. The contact planner provides geometrically precise, goal-dependent interaction targets, while RL synthesizes the robust motion required between contacts. Consequently, reference motions act solely as training guidance and are entirely eliminated at deployment, enabling a single onboard-sensing policy to autonomously execute and transition between diverse whole-body behaviors.

To support this framework, we developed two enabling pipelines. First, a constrained motion-retargeting tool extracts diverse contact strategies from human data, preserving key environmental interactions while respecting the robot's kinematic limits.  Second, because precise contact planning on sparse structures like ladder rungs demands high-fidelity perception, we introduce a LiDAR-based neural mapping pipeline that reconstructs detailed local geometry to provide the planner with a consistent terrain representation. Together, these components provide diverse contact-rich training data and sufficiently detailed terrain observations for precise whole-body contact planning in real-world environments.

\subsection*{Contact-Informed Whole-Body Tracker}
To provide whole-body reference motions for training, we utilize contact-aware motion retargeting~\cite{zewei26}. It transfers recorded human behaviors to the target humanoid while accounting for differences in morphology, kinematic constraints, and, importantly, body–environment interactions. Direct joint-angle mapping is generally infeasible because human motion data and the robot differ substantially in their kinematic structures and body proportions, often resulting in joint-limit violations, ground penetration, and physically inconsistent contacts. Such artifacts are particularly problematic for contact-rich behaviors such as climbing, crawling, vaulting, etc where the semantics of the motion are largely determined by which body part establishes contact, when the contact occurs, and where it occurs in the environment. We therefore explicitly extract the contact sequence together with the robot kinematics from the dataset.

This retargeting tool first converts human motions from different sources into a unified floating-base and Cartesian landmark representation, followed by spatial rescaling and temporal normalization. For each frame, it solves a nonlinear IK problem over the robot base pose and joint configuration. Position objectives track corresponding body landmarks, including the hips, knees, ankles, toes, shoulders, elbows, hands, chest, and head, while orientation objectives align individual robot limb segments with the corresponding directions in the human skeleton. The optimization additionally constrains the robot within its joint limits and enforces unilateral ground constraints on the feet to prevent penetration. For motions involving richer environmental interactions, minimum-distance constraints against the robot and environment geometry further prevent physically invalid intersections with obstacles such as boxes, stairs, and ladder rungs. Temporal regularization toward the previous solution suppresses frame-to-frame discontinuities, producing smooth trajectories while maintaining consistent contact configurations over time.

In parallel with kinematic retargeting, we explicitly detect and record body–environment contacts at every frame, involving the pelvis, knees, feet, elbows, and hands, which are identified from the simulated contact geometry and associated with the corresponding environmental objects. We record both the active contact sequence and contact locations in the world frame, preserving the temporal contact schedule and spatial relationship between the robot and the environment.  

Some motions in the training dataset were already retargeted to the Unitree G1 robot. For these motions, we only generate contact labels using this tool. Table \ref{data_source} summarizes the data sources and indicates, for each motion type, whether the data were retargeted and whether contact labels were generated using this retargeting tool.

After the motion dataset is prepared, we train a tracker to track the recorded contacts. The tracker is a three-layer multilayer perceptron with hidden dimensions $\{512,256,128\}$ and ELU activations; the network architectures are summarized in Table~\ref{tab:supp_networks}. The tracker receives a five-step proprioceptive history, a $9 \times 7$ local height scan, nine contact targets, and the goal command. Invalid contact targets are encoded as zero vectors. The policy outputs 23 joint-position offsets at $50\,\mathrm{Hz}$; these are scaled by $0.25$ and added to the default joint configuration to obtain targets for joint-level PD control at $200\,\mathrm{Hz}$. An overview of the tracker design is shown in Figure \ref{fig:overview}B. Figure~\ref{fig:ablation_v14}A shows that contact commands provide sufficient guidance for executing the demonstrated whole-body skills without additional reference kinematics as tracker inputs. As a result, the same tracker enables a broad range of behaviors, including sitting, ladder climbing, stair traversal, locomotion over sparse footholds, box climbing, vaulting, etc. Moreover, the contact-tracking framework captures fine-grained variations of motion on the same terrain by conditioning on different contact sequences, as discussed in \hyperref[sec:wbloco]{RESULTS A}. 

\subsection*{Attention-based Whole-Body Contact Planner}
At every control step, the planner outputs a contact plan for the nine contact-active bodies of the humanoid: the two feet, two knees, two hands, two elbows, and the pelvis, so that the plan can describe genuine multi-contact behavior rather than footstep sequences alone. For each body the plan gives a target contact position, expressed in a gravity-aligned frame centered at the robot base and rotated with its yaw so that it is invariant to the robot's world position and heading, together with a binary occupancy flag indicating whether that body is expected to establish a new contact within the planning horizon.

As illustrated in Figure \ref{fig:overview}C, the planner reads two streams. The first is a proprioceptive state vector including base angular velocity, projected gravity, joint positions, joint velocities, and previous actions, and a goal command encoded as a bounded horizontal direction vector and a relative heading. The second is an exteroceptive height scan, a $17 \times 11$ grid of three-dimensional terrain points sampled around the base in the body-yaw frame. Both tasks use the same three-dimensional goal-command format: a bounded horizontal direction vector and a relative heading. Reference-following derives this command from a future reference state, whereas goal-reaching derives it from the active terrain goal. The direction vector decreases in magnitude near the target.

The central design choice is that the planner \emph{selects} contacts from the perceived surface rather than regressing them in free three-dimensional space. A two-layer convolutional encoder with $5 \times 5$ kernels and zero padding extracts local terrain features, which are combined with per-point linear features to form the attention keys and values. The five-step proprioceptive history, goal command, and a 64-dimensional pooled terrain feature are encoded into a shared 96-dimensional feature vector. A learned embedding for each contact body generates channel-wise scale and bias parameters through feature-wise linear modulation (FiLM), transforming this shared vector into nine body-specific attention queries. A multi-head attention module (attention dimension $d=96$, $h=32$ heads) computes, for each body, a distribution of affordance weights over the height-scan points. The decoder combines the top-$k$ points using normalized, temperature-scaled weights. In the quantitative evaluations, we use $k=1$, so each retained contact target coincides with an observed terrain point. A classification head combines the body's attention output with pooled terrain context to predict contact occupancy. Network dimensions and decoding settings are listed in Tables~\ref{tab:supp_networks} and~\ref{tab:supp_replay}. 

Figure \ref{fig:generalization} visualizes the contact plans (shown as red dots) generated by the contact planner, which is trained in a supervised manner using rollout data collected from the whole-body tracking policy. The resulting contact plans demonstrate precise and versatile contact prediction across diverse terrains, without explicit skill labels. At each control step, the planner predicts the next contact location for each of nine body parts within a $1.5\,\mathrm{s}$ horizon. If no contact is predicted for a given body part within this horizon, the corresponding output is set to zero. The planner autonomously identifies appropriate contact locations and body parts for different terrain geometries, producing distinct contact strategies for behaviors such as climbing, vaulting, and sparse-terrain traversal. Importantly, contact planning is conditioned on the goal command, allowing the planner to generate different contact sequences for the same terrain according to the desired position and orientation. For example, as shown in Figure \ref{fig:generalization}A, the policy generates completely different behaviors on the same type of terrain given different local terrain geometries and different goal commands. Details are discussed in \hyperref[sec:wbloco]{RESULTS A}.


\subsection*{Three-Stage Training Pipeline}
The framework is trained in three sequential stages, each lifting a different reliance on supervision. The first stage trains the tracker alone on a curated reference-motion dataset covering climbing, stair traversal, walking on rough ground, and stepping-stone gaits, with the simulation terrain spatially aligned to each reference clip. The tracker receives ground-truth contact positions read directly from the clips in place of planner output, and episodes are initialized via Curriculum State Initialization (CSI; see Terrains and Curriculum). Supplying contacts as an oracle lets the tracker learn to execute them before any planner exists, yielding a tracker that has absorbed whole-body coordination priors --- including the non-trivial multi-limb sequencing that hand-crafted reward engineering struggles to elicit.

The second stage introduces the contact planner and trains it jointly with the tracker on the same reference-motion terrains. The planner is supervised on contact targets drawn from a replay buffer of successful rollouts and labeled by the hindsight procedure described in the next subsection. For each body $b$ with a valid upcoming contact $\mathbf{p}_b^{*}$ (indicated by the hindsight occupancy flag $c_b^{*}$), we convert the target into a soft distribution over the $LW = 187$ height-scan points ($L = 17$, $W = 11$) by placing an isotropic three-dimensional Gaussian of bandwidth $\sigma_a$ at $\mathbf{p}_b^{*}$ and normalizing:
\begin{equation}
p^{*}_{b,i} \;=\;
\frac{\exp\!\bigl(-\lVert \mathbf{h}_i - \mathbf{p}_b^{*}\rVert_2^{2}\,/\,2\sigma_a^{2}\bigr)}
     {\sum_{j=1}^{LW}\exp\!\bigl(-\lVert \mathbf{h}_j - \mathbf{p}_b^{*}\rVert_2^{2}\,/\,2\sigma_a^{2}\bigr)}.
\label{eq:soft_target}
\end{equation}
The planner outputs an affordance distribution $\alpha_{b,i}$ over the same points and a per-body occupancy probability $\hat{c}_b$, trained against this target with a masked cross-entropy plus a binary cross-entropy:
\begin{equation}
\begin{aligned}
\mathcal{L}_{\mathrm{planner}}
={}&-\frac{
\sum_{n=1}^{N}\sum_{b=1}^{9}c_{n,b}^{*}
\sum_{i=1}^{LW}p_{n,b,i}^{*}\log\alpha_{n,b,i}
}{
\max\!\left(1,\sum_{n=1}^{N}\sum_{b=1}^{9}c_{n,b}^{*}\right)
}\\
&-\frac{1}{9N}\sum_{n=1}^{N}\sum_{b=1}^{9}
\left[
c_{n,b}^{*}\log\hat{c}_{n,b}
+(1-c_{n,b}^{*})\log(1-\hat{c}_{n,b})
\right].
\end{aligned}
\label{eq:planner_loss}
\end{equation}
For a batch of $N$ samples, the affordance loss is averaged over valid upcoming contacts, while the occupancy loss is averaged over all nine bodies in every sample. A soft three-dimensional target, rather than a one-hot cell, is essential on thin supports such as ladder rungs, where the nearest cell in the horizontal plane can miss the structure entirely. We set $\sigma_a = 0.1\,\mathrm{m}$.

We gradually transfer contact guidance from the reference motion to the planner using separate input and reward schedules. At each control step, each environment independently receives planner-predicted contacts with probability $\eta_{\mathrm{obs}}$ and reference contacts otherwise. A later-starting schedule, $\eta_{\mathrm{rew}}$, reduces the reference-contact reward weights and increases the planner-contact reward weight, as specified in Table~\ref{tab:rew_func}. This transition trains the tracker to follow the planner's terrain-adapted contacts while retaining the remaining reference-motion guidance. Both schedules are defined in \eqref{eq:supp_contact_annealing}.

The third stage, as illustrated in Figure \ref{fig:stage3}, takes the system beyond the demonstration distribution through a bootstrapping loop between the contact planner and the whole-body controller, similar to~\cite{PARC, zewei26}. We augment training with procedurally generated \emph{goal-reaching} fine-tuning terrains for which no reference motions are available, including platforms with randomized orientation, stepping stones, gaps, and pallets of varying width, balance beams, climbable obstacles, and ladders of varying rung spacing, etc. On these goal-reaching terrains, the contact planner first proposes contact plans conditioned on the local map, robot proprioception, and goal command. The whole-body controller then executes these plans and is fine-tuned with a contact-tracking reward that encourages it to realize the planner-predicted contacts while the task reward (Table~\ref{tab:rew_func}) drives the robot toward a goal sampled on a reachable terrain patch. Episodes have a maximum duration of $15\,\mathrm{s}$ and may terminate earlier under the configured failure conditions. Successful rollouts are added to the planner's hindsight buffer, which in turn provides new supervision for the contact planner. The planner is therefore refined using contact strategies that were successfully realized by the controller on terrains not covered by the original demonstrations, and its improved predictions generate progressively better training targets for subsequent controller rollouts. Through this mechanism, the learned contact strategies generalize to terrains outside the original motion dataset. In parallel, reference-trajectory training is retained through proportional sampling from the original dataset to preserve previously learned tracking behaviors. To account for the different reward scales, we use separate value networks and normalize advantages separately for reference-following and goal-reaching environments within each PPO update.

Figure \ref{fig:generalization}B demonstrates the fine-tuned policy on a selection of terrains unseen during training, including novel compositions of training terrain primitives and scanned meshes that are absent from the original tracking dataset used in the preceding training stages. Starting from the learned contact planner and whole-body contact tracker, we fine-tune the policy on procedurally altered variants of the original terrains, with randomized obstacle poses, geometries, and configurations. Through this process, the policy learns to adapt both its contact strategy and whole-body motion to terrain variations while retaining the behaviors acquired from the original motion data.

\subsection*{Hindsight Contact Labeling}
The planner's targets come from the policy's own successful experience rather than from the motion dataset, so that it learns contacts the tracker can actually realize. After each completed episode we scan its recorded trajectory backward to identify the next contact onset for each body part. We identify contact onset from the vertical contact force after discarding contact segments shorter than ten consecutive control steps, thereby filtering out brief or intermittent contacts. For each timestep, the next contact position within the $1.5\,\mathrm{s}$ horizon is transformed into that timestep's body-yaw frame and retained as a target if it lies within the local scan footprint. Only successful episodes enter the replay buffer (as shown in Figure \ref{fig:stage3}): for reference-following environments, those initialized within the first $10\%$ of the reference motion and progressing to at least $90\%$ of its duration; for goal-reaching environments, those that at some point bring the robot within $0.5\,\mathrm{m}$ of the goal horizontally while within $0.1\,\mathrm{m}$ of it vertically. For goal-reaching terrains, the buffer is partitioned by terrain tile, with quality-based eviction and balanced sampling across terrain types and tiles. The quality metric favors lower positive mechanical work per unit initial horizontal goal distance. Following planner initialization, optimization is decoupled from the on-policy loop: we draw from the replay buffer for ten planner optimizer updates per PPO iteration. The labeling thresholds, replay-sampling settings, and optimizer parameters are listed in Table~\ref{tab:supp_replay}.

\subsection*{LiDAR-based Neural Mapping}
\label{sec:neumap}

\begin{figure*}
    \centering
    \includegraphics[width=1\linewidth]{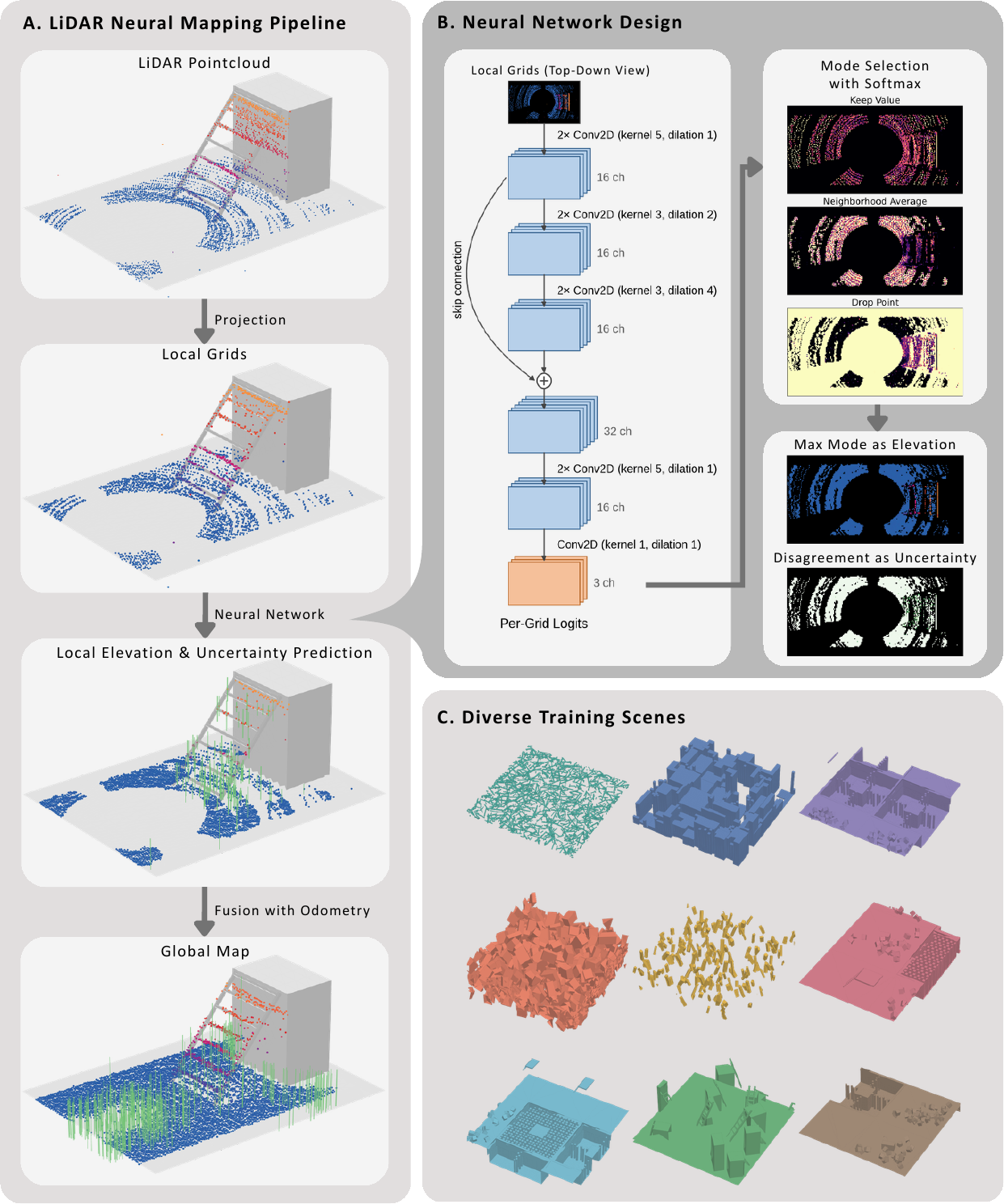}
    \caption{Overview of the LiDAR-based neural mapping pipeline for height map observations. (A) At each frame, LiDAR points are projected into x-y grid cells and the highest point per cell is retained, producing a 2D local grid map. A neural network then predicts local elevations with associated uncertainty, and the predictions are fused into the global map using odometry. (B) The network uses CNNs to predict per-grid logits over three fill modes: the exact measured value, the neighborhood average, or leaving the grid empty. The mode with the highest logit determines the grid's elevation prediction, while the disagreement among modes and logits yields the uncertainty estimate. (C) The network is trained on synthetic data generated from diverse scene meshes.}
    \label{fig:neumap}
\end{figure*}

To provide the controller with accurate height map observations, we design a
LiDAR-based neural mapping pipeline, illustrated in
Figure~\ref{fig:neumap}. The pipeline builds on the depth-based neural
mapping method of Zhang et al.~\cite{ame2}, adapted to the sensing characteristics of
LiDAR to handle perception noise. We convert each frame's point cloud into
local grids and use a neural network to predict local elevations with
associated uncertainty. The predictions are fused into a global map, which
the locomotion controller queries during deployment.

\paragraph{Network design.}
The main difference from Zhang et al.~\cite{ame2} lies in the network design. Unlike
depth cameras, which provide a roughly fixed region of interest, our
head-mounted MID-360~\cite{livox_mid360} LiDAR provides a panoramic view
with sparse hits. Rather than directly predicting a dense reconstruction,
we have the network select one of three modes for each grid:
keep the exact measured height, use the average of the neighborhood, or
discard the value and leave the grid empty. This allows the model to
generalize to arbitrary scan sparsity. Moreover, the disagreement among
the mode logits, interpreted as probabilities for sampling modes, serves
as an uncertainty estimate that accounts for sensing noise. Only non-empty
predictions are processed during fusion.

\paragraph{Fusion prior for overhanging structures.}
Since the mapping must handle overhanging structures such as ladders, we
introduce a heuristic prior in the global fusion. When a new prediction falls in the
same grid as an existing confident prediction (uncertainty $<0.15$ m) and lies more than 0.15~m
below it, the new point is discarded unless its elevation angle is above
$-40^{\circ}$. This ensures that if an earlier scan observes an
overhanging ladder rung while a later close-by scan only observes the floor beneath
it, the rung is not erased from the global map. Despite this ladder-targeted prior, the mapping module stays effective for all tested terrains.

\paragraph{Training and implementation.}
We train on the same meshes as Zhang et al.~\cite{ame2}, augmented with random
overhanging bars and random ladders leaning against boxes to ensure
performance on sparse overhanging structures. We also simulate per-point
positional noise and ray-divergence
noise~\cite{ongan2026learningagileperceptivetraversal} according to the
LiDAR specifications. Since training is formulated as mode
classification, we derive labels from the ground truth and the generated
noisy grids, and train the CNN with a cross-entropy loss. Local grids
cover $[-3, 3]$~m in $x$ and $[-1.5, 1.5]$~m in $y$ in the gravity-aligned LiDAR frame,
with a grid resolution of 3~cm. The pipeline runs onboard on the
Jetson Orin NX CPU, updating within 30~ms per 10-Hz LiDAR scan (20000 points).
It is also worth noting that, since the LiDAR is head-mounted, occlusion is
considerably less severe than on the platforms used by Zhang et al.~\cite{ame2}. This allows
us to obtain a high-quality map observation directly by inpainting
high-uncertainty regions with the neighborhood minimum followed by a
median filter. 

\subsection*{Observation Space}
We use an asymmetric actor--critic design for tracker training: the actor observes only quantities available on the real robot, while the critics may use privileged simulation state. Across all three stages, the actor receives a five-step proprioceptive history, a $9 \times 7$ local height scan, contact targets for nine bodies, and the goal command $\mathbf{g}$. Stage~1 uses reference-derived contact targets. Stage~2 gradually replaces these targets with planner predictions through stochastic selection per environment, and Stage~3 uses planner predictions throughout training. Predicted targets are retained when their occupancy probability exceeds $0.5$ and are otherwise set to zero. The scan spacing, spatial coverage, and goal-command parameters are listed in Table~\ref{tab:supp_observation_curriculum}.

Two separate value networks, $V_m$ and $V_{\mathrm{fs}}$, value the reference-following and goal-reaching regimes; both see privileged information withheld from the actor. The reference-following critic $V_m$, used in Stages~1--2 and on reference-anchored terrains in Stage~3, receives the current proprioceptive state without added observation noise or temporal stacking, the full $17 \times 11$ height scan, and the base linear velocity. To support value estimation during reference-following, it additionally receives three privileged streams from the reference motion, none exposed to the actor: a $52$-dimensional reference kinematic vector containing base linear and angular velocities, joint positions, and joint velocities, the $27$-dimensional reference contact-body positions $\mathbf{p}_b^{\mathrm{ref}}$, and the $9$-dimensional reference contact flags $c_b^{\mathrm{ref}}$.

Stage~3 introduces the goal-reaching critic $V_{\mathrm{fs}}$, where no reference motion exists. It keeps the current proprioceptive state, base linear velocity, and full height scan, but replaces the reference streams with the goal command $\mathbf{g}$, horizontal distance to the goal $d_g$, vertical displacement to the target base height $\Delta z = z_{\mathrm{goal}} - z_{\mathrm{base}}$, elapsed episode time normalized by the episode horizon, and the planner's predicted contacts $\hat{\mathbf{p}}_b$. A dispatcher routes each environment to $V_m$ or $V_{\mathrm{fs}}$ by its type. Table~\ref{tab:symbol} summarizes the notation.


\begin{table}[h]
\centering
\caption{\textbf{Nomenclature.} Reward kernel bandwidths $\sigma_{(\cdot)}$ and other hyperparameter values are introduced inline where used.}
\label{tab:symbol}

\resizebox{\columnwidth}{!}{%
\begin{tabular}{cl}
\hline
$\mathbf{r}_b, \mathbf{v}_b, \boldsymbol{\omega}_b$ & base position, linear and angular velocity \\
$\theta_e$ & angular difference between actual and reference base orientations \\
$\mathbf{q}_j, \dot{\mathbf{q}}_j$ & joint position and velocity ($_0$: default) \\
$\boldsymbol{\tau}_j$ & joint torque \\
$\mathbf{a}_t$ & action at control step $t$ \\
$(\cdot)^{*}$ & supervision target, defined by context \\
$\hat{\mathbf{p}}_b, \hat{c}_b$ & planner-predicted contact position and occupancy probability \\
$\mathbf{p}_b^{*}, c_b^{*}$ & hindsight contact target and occupancy flag \\
$\mathbf{p}_b^{\mathrm{ref}}, c_b^{\mathrm{ref}}$ & instantaneous reference contact-body positions and contact flags \\
$\mathbf{p}_b^{\mathrm{exec}}$ & executed (actual) contact-body position \\
$c_b^{\mathrm{meas}}$ & measured contact flag, defined in Table~\ref{tab:rew_func} \\
$\alpha_{b,i}$ & affordance weight on height-scan point $i$ for body $b$ \\
$\eta_{\mathrm{obs}}$ & probability of using planner-predicted contact inputs during Stage~2 \\
$\eta_{\mathrm{rew}}$ & transition coefficient for Stage~2 contact rewards \\
$\mathbf{h}_i$ & $i$-th height-scan point in body-yaw frame \\
$\mathbf{g}$ & bounded horizontal goal direction and relative heading in the body-yaw frame \\
$d_g, \Delta z$ & horizontal distance and signed vertical offset to goal \\
$\mathbf{F}_f, \mathbf{v}_f, c_f$ & contact force, velocity, and binary state of contact link $f$ \\
$V_m, V_{\mathrm{fs}}$ & mimic and goal-reaching critic value functions \\
\hline
\end{tabular}%
}

\end{table}

\subsection*{Reward Functions}
The per-step reward is a weighted sum of task, regularization, and style terms (Table~\ref{tab:rew_func}). Task terms drive imitation of the reference motion and, in Stage~3, goal reaching on procedural terrains; regularization terms penalize unphysical or unsafe joint motion; style terms shape the spatial and temporal regularity of the motion. The active subset and its weights vary by stage. Stage~1 supervises the tracker with ground-truth contact and reference-proprioception targets from the motion-capture corpus, alongside the full set of regularization terms, with all goal-reaching terms inactive. In Stage~2, the reference-contact reward weights are multiplied by $1-\eta_{\mathrm{rew}}$ and the planner-contact reward weight by $\eta_{\mathrm{rew}}$, following \eqref{eq:supp_contact_annealing}; all other weights remain unchanged. Stage~3 masks the mimic tracking terms to the reference-motion environments and activates, on goal-reaching environments, the goal-progress and stand-at-goal rewards together with goal-reaching-specific style terms (marked $^\dagger$).

\begin{table*}[h]
\centering
\caption{\textbf{Definition of reward terms across training stages.}
The per-step reward is $\Delta t\sum_i w_i r_i$, where $\Delta t$
is the control timestep and $w_i$ are the weights listed below.
Grey cells denote inactive terms.
Reference-tracking terms apply only to reference-anchored environments;
terms marked $^\dagger$ apply only to goal-reaching environments.
The Stage~2 contact weights use $\eta_{\mathrm{rew}}=\eta_{\mathrm{rew}}(u)$,
the reward-annealing coefficient defined in \eqref{eq:supp_contact_annealing}.
The index $b$ runs over the nine contact bodies, $j$ over the
23 controlled joints, and $f$ over monitored contact-sensor links.
Superscripts $\mathrm{ref}$ denote instantaneous reference quantities;
$B$ and $W$ denote body-yaw and world coordinates.
Actual and reference body-yaw positions use their respective base poses.
$\theta_e$ is the angular difference between actual and reference
base orientations,
$\chi_{\mathrm{mov}}=\mathds{1}[\|\mathbf{v}_b^{*}\|>0.1]$,
and $m_b$ indicates a nonzero, occupancy-gated planner target.
$c_b^{\mathrm{meas}}$ indicates a vertical contact-force magnitude
exceeding $1\,\mathrm{N}$ within the sensor-history window.
$\operatorname{wrap}_{\pi}$ wraps angular differences to $[-\pi,\pi)$.
The kernel scales are $\sigma_p=0.5$, $\sigma_r=1$,
$\sigma_v=1$, $\sigma_\omega=3$, $\sigma_q=2$,
$\sigma_{\dot q}=10$, and $\sigma_c=1$, in the corresponding SI units.
For goal rewards,
$g_{xy}=\mathds{1}[d_g<0.5]$,
$g_h=\mathds{1}[|\Delta z|<0.2]$,
$g_\psi=\mathds{1}[|\Delta\psi|<0.5]$, and
$g_v=\mathds{1}[v_{\to g}>0.1]\mathds{1}[\|\mathbf{v}_{b,xy}^{W}\|\leq v_{\max}]$.
Here, $d_g$ is the horizontal goal distance,
$\Delta z$ is the target base height minus the current base height,
$\Delta\psi$ is the target-heading error, and $v_{\to g}$ is the
horizontal velocity projected toward the goal.
The speed cap is $0.8\,\mathrm{m\,s^{-1}}$ on the configured stair
terrains and $2.0\,\mathrm{m\,s^{-1}}$ elsewhere.
For get-up terrains,
$g_z=\mathds{1}[v_{b,z}^{W}\operatorname{sign}(\Delta z)>0.1
\lor |\Delta z|<0.2]$; otherwise, $g_z=1$.
$d_{\mathrm{foot}}$ is the fraction of feet without measured contact,
$d_{\mathrm{tilt}}=1-(g_z^{\mathrm{base}})^2$ uses the vertical
component of projected gravity, and
$d_q=\frac{1}{23}\sum_j|q_j-q_{0,j}|$.
$T_i$ denotes the terrain mask for style term $i$.
$F_{f,xy}^{\max}$ and $F_{f,z}^{\max}$ are the maximum horizontal
force magnitude and vertical force over the sensor-history window;
$t_f^{\mathrm{contact}}$ is the continuous contact duration.
$\mathbf{g}_{\mathrm{torso}}$ is gravity projected into the torso frame.
Computed and applied torques are taken before and after actuator
limiting, respectively.
$[x]_+=\max(x,0)$, and joint bounds are the configured soft limits.
All thresholds use SI units.}
\label{tab:rew_func}

\begin{tabular}{|c|l|c|c|c|c|}
\hline
Category & Term & Functional Form & Stage 1 & Stage 2 & Stage 3 \\
\hline

\multirow{8}{*}{Task}
& track\_base\_pos
& $\exp\!\left(
-\chi_{\mathrm{mov}}\|\mathbf{r}_b-\mathbf{r}_b^{*}\|^2/\sigma_p^2
-\theta_e^2/\sigma_r^2
\right)$
& $3.0$ & $3.0$ & $3.0$ \\

& track\_base\_vel
& $\exp\!\left(
-\|\mathbf{v}_b-\mathbf{v}_b^{*}\|^2/\sigma_v^2
-\|\boldsymbol{\omega}_b-\boldsymbol{\omega}_b^{*}\|^2/\sigma_\omega^2
\right)$
& $1.0$ & $1.0$ & $1.0$ \\

& track\_joint\_pos
& $\exp\!\left(
-\|\operatorname{wrap}_{\pi}(\mathbf{q}_j-\mathbf{q}_j^{*})\|^2/\sigma_q^2
\right)$
& $2.5$ & $2.5$ & $2.5$ \\

& track\_joint\_vel
& $\exp\!\left(
-\|\dot{\mathbf{q}}_j-\dot{\mathbf{q}}_j^{*}\|^2/\sigma_{\dot q}^2
\right)$
& $0.5$ & $0.5$ & $0.5$ \\

& track\_contact\_label
& $\frac{1}{9}\sum_b
c_b^{\mathrm{meas}}(2c_b^{\mathrm{ref}}-1)$
& $5.0$
& $5.0(1-\eta_{\mathrm{rew}})$
& \cellcolor[HTML]{9B9B9B} \\

& track\_contact\_pos\_base
& $\exp\!\left(
-\sum_b\|\mathbf{p}_b^{\mathrm{exec},B}
-\mathbf{p}_b^{\mathrm{ref},B}\|^2/\sigma_c^2
\right)$
& $2.0$
& $2.0(1-\eta_{\mathrm{rew}})$
& \cellcolor[HTML]{9B9B9B} \\

& track\_contact\_pos\_global
& $\exp\!\left(
-\chi_{\mathrm{mov}}\sum_b
\|\mathbf{p}_b^{\mathrm{exec},W}
-\mathbf{p}_b^{\mathrm{ref},W}\|^2/\sigma_c^2
\right)$
& $1.0$
& $1.0(1-\eta_{\mathrm{rew}})$
& \cellcolor[HTML]{9B9B9B} \\

& track\_planned\_contact
& $\exp\!\left(
-\frac{\sum_b m_b
\|\mathbf{p}_b^{\mathrm{exec},B}-\hat{\mathbf{p}}_b\|^2}
{\sigma_c^2\max(1,\sum_b m_b)}
\right)$
& \cellcolor[HTML]{9B9B9B}
& $1.5\eta_{\mathrm{rew}}$
& $1.5$ \\

\hline

\multirow{2}{*}{Task$^\dagger$}
& move2goal
& $(1-g_{xy})g_v+g_{xy}g_z$
& \cellcolor[HTML]{9B9B9B}
& \cellcolor[HTML]{9B9B9B}
& $40$ \\

& standatgoal
& $g_{xy}g_hg_\psi
\exp\!\left(-(d_{\mathrm{foot}}+d_{\mathrm{tilt}}+d_q)/3\right)$
& \cellcolor[HTML]{9B9B9B}
& \cellcolor[HTML]{9B9B9B}
& $20$ \\

\hline

\multirow{6}{*}{Regularization}
& action\_rate
& $\|\mathbf{a}_t-\mathbf{a}_{t-1}\|^2$
& $-0.1$ & $-0.1$ & $-0.1$ \\

& action\_rate\_2nd
& $\|\mathbf{a}_t-2\mathbf{a}_{t-1}+\mathbf{a}_{t-2}\|^2$
& $-5{\times}10^{-2}$
& $-5{\times}10^{-2}$
& $-5{\times}10^{-2}$ \\

& joint\_torque
& $\sum_j(\tau_j^{\mathrm{applied}})^2$
& $-1{\times}10^{-5}$
& $-1{\times}10^{-5}$
& $-1{\times}10^{-5}$ \\

& torque\_limits
& $\sum_j|\tau_j^{\mathrm{computed}}-\tau_j^{\mathrm{applied}}|$
& $-1{\times}10^{-3}$
& $-1{\times}10^{-3}$
& $-1{\times}10^{-3}$ \\

& dof\_pos\_limits
& $\min\!\left(
30,\sum_j([q_j^{\min}-q_j]_+ + [q_j-q_j^{\max}]_+)
\right)$
& $-10$ & $-10$ & $-10$ \\

& dof\_vel\_limits
& $\sum_j\min\!\left(
1,[|\dot q_j|-0.9\dot q_j^{\max}]_+
\right)$
& $-1$ & $-1$ & $-1$ \\

\hline

Style
& contact\_acc
& $\min\!\left(
10^4,\sum_b\|\ddot{\mathbf{p}}_b^{\mathrm{exec},W}\|
\right)$
& $-0.01$ & $-0.01$ & $-0.01$ \\

\hline

\multirow{4}{*}{Style$^\dagger$}
& flat\_orientation
& $T_{\mathrm{ori}}\|\mathbf{g}_{\mathrm{torso},xy}\|^2$
& \cellcolor[HTML]{9B9B9B}
& \cellcolor[HTML]{9B9B9B}
& $-10$ \\

& stumbling
& $T_{\mathrm{st}}\sum_f
\mathds{1}[F_{f,xy}^{\max}>1.1F_{f,z}^{\max}+1]$
& \cellcolor[HTML]{9B9B9B}
& \cellcolor[HTML]{9B9B9B}
& $-10$ \\

& slippage
& $T_{\mathrm{sl}}\sum_f
\mathds{1}[t_f^{\mathrm{contact}}>0.026]\,
\mathds{1}[\|\mathbf{v}_f\|>0.5]$
& \cellcolor[HTML]{9B9B9B}
& \cellcolor[HTML]{9B9B9B}
& $-10$ \\

& ang\_vel\_xy
& $T_{\omega}\|\boldsymbol{\omega}_{\mathrm{torso},xy}^{W}\|^2$
& \cellcolor[HTML]{9B9B9B}
& \cellcolor[HTML]{9B9B9B}
& $-0.1$ \\

\hline
\end{tabular}
\end{table*}

\subsection*{Training}
We train all policies in Isaac Lab~\cite{isaac_lab}. The contact tracker is trained using a customized Proximal Policy Optimization (PPO) algorithm~\cite{ppo}, while the contact planner is trained with supervised learning on rollout data collected from the tracker. The training pipeline consists of three stages. Stage 1 requires approximately 1 day using 3 NVIDIA RTX 4090 GPUs. Stage 2 requires approximately 1 day and 21 hours using 2 RTX 4090 GPUs, and Stage 3 requires approximately 2 days and 6 hours using 4 RTX 4090 GPUs. Planner optimization settings are listed in Table~\ref{tab:supp_replay}, and PPO settings are listed in Table~\ref{tab:ppo}. 

\paragraph{Domain Randomization}
At simulation startup we sample, independently per environment, the friction and restitution coefficients, torso mass and center-of-mass offset, joint stiffness and damping scales, and per-joint default-position offsets, from uniform ranges listed in the supplementary methods; each environment retains its sampled values throughout training. At every control step, uniform noise is added to all non-privileged (actor) observations; the height scan additionally receives Gaussian noise per step and a per-episode elevation offset to emulate elevation-mapping drift. These randomizations are active across all three training stages. Details are summarized in Table \ref{tab:domain_randomization}.

\paragraph{Terrains and Curriculum}
Stages 1 and 2 use \emph{reference-anchored} terrains spatially aligned with the motion-capture clips (stairs of varying step height, gaps, stepping stones, climbing obstacles, balance beams). In Stage~1 the tracker is initialized by Curriculum State Initialization (CSI): each episode begins at a curriculum-set phase of the reference motion, starting near the end of each clip (initial phase $0.9$) and widening backward toward the start as tracking improves --- so the tracker first learns to finish each clip before being asked to execute it from the beginning. Stage~2 disables this phase curriculum and samples initial phases uniformly over each clip. Stage~3 adds \emph{goal-reaching} terrains from a procedurally generated suite (platforms, stepping stones, gaps, balance beams, ladders, pallets), each paired with reachable target patches sampled at reset. These terrains span ten difficulty levels set by step height, gap width, or stone spacing; following the curriculum of~\cite{Rudin2021}, an environment is promoted a level when it reaches within $0.5\,\mathrm{m}$ horizontally of its goal at the required height and completes the episode without early termination (gated by an exponentially smoothed per-environment success rate) and demoted when its minimum qualifying goal distance during the episode exceeds $1.0\,\mathrm{m}$. The initialization and curriculum-update parameters are summarized in Table~\ref{tab:supp_observation_curriculum}.

\subsection*{Deployment}
We deploy the trained generalist policy on the Unitree G1 using its onboard NVIDIA Jetson Orin, with all perception, state estimation, mapping, and policy inference running entirely onboard. The G1's Livox LiDAR provides point-cloud measurements to DLIO~\cite{chen2022dlio}, which estimates the robot root state and produces a deskewed point cloud. The deskewed point cloud is subsequently processed by our neural mapping pipeline to generate the local height scan used as terrain observation by the policy. The neural mapping module, DLIO, and the learned policy run concurrently on the Jetson Orin, enabling fully onboard perception and control without external computation or sensing.



\bibliography{scibib}

\section*{Acknowledgments}
We thank Ryan Batke, Ioannis Dadiotis, Emre Elbir, Yuni Fuchioka, Elena Krasnova, Chenhao Li, Adria Lopez, Roman Peneder, Iuri Pereira Barros, and Thiago Henrique Segreto Silva for their support with infrastructure, video recording, and hardware testing, Shengcheng Fu for providing locomotion policies for data collection, Francesca Bray and Changan Chen for mocap data collection, and Unitree for providing support on hardware.   
\textbf{Funding:}
This work was funded in part by the NCCR Automation, ETH RobotX, EU Project 101121321, the ETH Mobility Initiative, Armasuisse W\&T, ETH AI Center, Fourier Intelligence, and Apple Inc. Any views, opinions, findings, and conclusions or recommendations expressed in this material are those of the authors and should not be interpreted as reflecting the views, policies or position, either expressed or implied, of Apple Inc. 
\textbf{Author contributions:} 
 J.~H., K.~W. and C.~Z. formulated the main ideas, trained the controllers, and conducted most of the experiments. Z.~Z., C.~S., and K.~Q. helped with experiments and provided insights into system development. V.~K. and M.~H. acquired the funding for the project. All authors helped to write, improve, and refine the paper.  

\clearpage

\section*{\highlight{Supplementary Methods}}
\setcounter{figure}{0}
\renewcommand\thefigure{S\arabic{figure}}
\setcounter{table}{0}
\renewcommand\thetable{S\arabic{table}}


\subsection*{Training Details}

\subsubsection*{Data source}
Source of the motion data is listed in Table \ref{data_source}. 
\begin{table}[]
\caption{Source of the motion data. Retargeted: retargeted using our method; Contact labeled: contact labeled with our method.}
\begin{tabular}{l|l|l|l}
                   & Data origin  & Retargeted & Contact labeled \\ \hline
Box climbing       & Omniretarget \cite{yang2026omni} & No                          & Yes                              \\ \hline
Stairs climbing    & Mixamo \cite{Mixamo}       & Yes                         & Yes                              \\ \hline
Ladder climbing    & Mixamo \cite{Mixamo}      & Yes                         & Yes                              \\ \hline
Sit down           & Mixamo \cite{Mixamo}      & Yes                         & Yes                              \\ \hline
Walk               & LAFAN \cite{harvey2020robust}       & Yes                         & Yes                              \\ \hline
Get up & LAFAN \cite{harvey2020robust}       & Yes                         & Yes                              \\ \hline
Vault              & Kimodo \cite{rempe2026kimodo}       & No                          & Yes                              \\ \hline
Sparse terrains    & AME policy \cite{fu2026global}   & No                          & No                              
\end{tabular}
\label{data_source}
\end{table}

\subsubsection*{Domain Randomizations}

Details of domain randomizations are summarized in Table \ref{tab:domain_randomization}.

\begin{table*}[!htbp]
\centering
\caption{\textbf{Domain randomization settings.}
$\mathcal{U}(a,b)$ denotes a uniform distribution over $[a,b]$.
Multiplicative factors are applied to nominal values.
Dynamics parameters are sampled once per environment at simulation start.}
\label{tab:domain_randomization}
\setlength{\tabcolsep}{5pt}
\renewcommand{\arraystretch}{1.1}
\begin{tabularx}{\textwidth}{
    >{\raggedright\arraybackslash}p{0.42\textwidth}|
    >{\raggedright\arraybackslash}X
}
Parameter & Value \\ \hline

\multicolumn{2}{l}{\textit{Dynamics}} \\
Contact friction (static/dynamic)
& $\mathcal{U}(0.3,1.2)$ / $\mathcal{U}(0.3,1.2)$; ground friction $=1.0$ \\
Restitution
& $\mathcal{U}(0.0,0.5)$ \\
Torso added mass
& $\mathcal{U}(-2.0,4.0)\,\mathrm{kg}$ \\
Leg link mass
& $\times\mathcal{U}(0.85,1.15)$ \\
Torso CoM offset
& $x:\pm0.025\,\mathrm{m}$, $y,z:\pm0.05\,\mathrm{m}$ \\
Joint zero-position bias (non-ankle)
& $\mathcal{U}(-0.01,0.01)\,\mathrm{rad}$ \\
Joint zero-position bias (ankle)
& $\mathcal{U}(-0.03,0.03)\,\mathrm{rad}$ \\
Joint armature
& $\times\mathcal{U}(0.9,1.1)$ \\
PD gains ($k_p,k_d$)
& $\times\mathcal{U}(0.9,1.1)$ \\
Delay
& $\mathcal{U}(0, 20)$ ms \\
\hline

\multicolumn{2}{l}{\textit{Proprioception observations}} \\
Base angular velocity
& $\mathcal{U}(-0.05,\,0.05)\,\mathrm{rad/s}$ \\
Projected gravity
& $\mathcal{U}(-0.0125,\,0.0125)$ \\
Joint positions
& $\mathcal{U}(-0.01,\,0.01)\,\mathrm{rad}$ \\
Joint velocities
& $\mathcal{U}(-1.5,\,1.5)\,\mathrm{rad/s}$ \\
Goal contact positions
& $\mathcal{U}(-0.1,\,0.1)\,\mathrm{m}$ \\
Goal direction / yaw
& $\mathcal{U}(-0.2,\,0.2)\,\mathrm{rad}$\\
\hline

\multicolumn{2}{l}{\textit{Exteroception}} \\
Planner height scan ($17\times11$)
& $\mathcal{N}(0,0.03^2)\,\mathrm{m}$\\
Tracker height scan ($9\times7$)
& $\mathcal{U}(-0.1,0.1)\,\mathrm{m}$ \\
Odometry delay
& $\mathcal{U}(50,100)\,\mathrm{ms}$ \\
Odometry XY drift
& Uniform within a $0.05\,\mathrm{m}$ radius disk \\
Odometry Z drift
& $\mathcal{U}(-0.05,0.05)\,\mathrm{m}$ \\
\hline

\multicolumn{2}{l}{\textit{Initial states}} \\
Reset joint position
& $\mathcal{U}(-0.2,0.2)\,\mathrm{rad}$ around reference/default pose \\
Reset base height
& $+\mathcal{U}(0,0.05)\,\mathrm{m}$ \\
Reset yaw (goal-reaching)
& $\mathcal{U}(-0.4,0.4)\,\mathrm{rad}$ \\
Initial phase (reference-following)
& Stage 1: curriculum;
Stage 2: $\mathcal{U}(0,0.98)$;
Stage 3: $0$ \\
\end{tabularx}
\end{table*}

\subsubsection*{Terrains}
\phantomsection
\label{sec:supp_training_terrains}
Reference-anchored terrains : 
\begin{itemize}
    \item \textbf{Dense foot }
    \begin{itemize}
        \item Walking and turning
        \item Stair ascent and descent
    \end{itemize}
    \item \textbf{Sparse foot}
    \begin{itemize}
        \item Beams - narrow, uninterrupted walking surfaces
        \item Grid and paired stepping stones - small, separated footholds arranged in grids or pairs
        \item Pallets - separated narrow slats
        \item Stakes - small pillar-shaped supports
    \end{itemize}
    \item \textbf{Dense whole-body}
    \begin{itemize}
        \item Box ascent and descent
        \item Sitting and standing transitions
        \item Getting up from face-up and face-down positions
    \end{itemize}
    \item \textbf{Sparse whole-body}
    \begin{itemize}
        \item Ladders (50$^\circ$ and 60$^\circ$ inclinations; 25cm ladder rung distance; 1m ladder rung length)
        \item Vaults (2m width; 0.2m thickness; 0.5m height) - crossing obstacles with hand support
    \end{itemize}
\end{itemize} 
\noindent Goal-reaching terrains:
\begin{itemize}
\item \textbf{Dense foot}
\begin{itemize}
\item Stair ascent and descent
\item Walled stairs - stairs bounded by side walls
\item Walk-up platforms
\item Rough ground and slopes

\end{itemize}
\item \textbf{Sparse foot}
\begin{itemize}
\item Gaps
\item Balance beams - narrow walking surfaces spanning gaps
\item Pallets - separated narrow slats
\item Double stones - paired stepping-stone supports
\item Discrete platforms - separated platforms with varying heights

\end{itemize}
\item \textbf{Dense whole-body}
\begin{itemize}
\item Box ascent and descent
\item Getting up from face-up and face-down positions
\item Uneven climbing terrain - climbing obstacles with irregular surfaces
\end{itemize}
\item \textbf{Sparse whole-body}
\begin{itemize}
\item Ladders with varied inclinations (50$^\circ$ - 65$^\circ$) and rung configurations (20-30cm rung distance range and 1-2m length range)
\item Varied vaults  (1.2-2m width range; 0.16-0.22m thickness range; 0.47-0.53m height range; 5$^\circ$ max roll; 3$^\circ$ max pitch; 10$^\circ$ max yaw)- crossing obstacles with hand support
\end{itemize}
\end{itemize}

\subsubsection*{Network architectures}

The tracker and both value networks use three hidden layers with
512, 256, and 128 units and ELU activations.
The contact-conditioned tracker receives 468 features:
375 proprioceptive-history features, 63 height samples,
27 contact-target coordinates, and a three-dimensional goal command.
The kinematics-conditioned variant adds 52 reference features.
Both trackers output commands for 23 joints.
The reference-following and goal-reaching critics receive
353 and 298 features, respectively.

Table~\ref{tab:supp_networks} specifies the planner architectures.
The attention planner uses ELU activations in its terrain encoder
and ReLU activations in its context, FiLM, and occupancy MLPs.
The MLP baseline uses two 285-unit hidden layers with ELU activations,
followed by separate terrain-weight and occupancy heads.
Terrain weights are normalized separately for each body,
and occupancy outputs are passed through a sigmoid.
The attention and MLP planners contain 786,993 and 833,322
parameters, respectively.

\begin{table*}[t]
\centering
\caption{\textbf{Contact-planner architectures.}
Both planners receive a five-step proprioceptive history,
a three-dimensional goal command, and 187 terrain points.
Arrows indicate layer widths; height-map encoder widths denote channels.}
\label{tab:supp_networks}
\renewcommand{\arraystretch}{1.15}
\begin{tabularx}{\textwidth}{@{}>{\raggedright\arraybackslash}p{0.33\textwidth}|>{\raggedright\arraybackslash}p{0.31\textwidth}|>{\raggedright\arraybackslash}X@{}}
Component & Dimensions & Details \\ \hline
\multicolumn{3}{@{}l@{}}{\textit{Attention planner (786,993 parameters)}} \\
Terrain-point embedding
& $3\rightarrow16$
& Linear projection; ELU \\
Height-map encoder
& $1\rightarrow16\rightarrow48$
& $5\times5$ kernels; padding 2; ELU \\
Terrain-feature projection
& $64\rightarrow96$
& Linear projection \\
Terrain-context encoder
& $96\rightarrow64\rightarrow64$
& Pointwise MLP; max pooling \\
Proprioception--goal--context encoder
& $442\rightarrow128\rightarrow128\rightarrow96$
& MLP; ReLU \\
Body embeddings
& $9\times96$
& Nine learned body embeddings \\
FiLM parameter generator
& $96\rightarrow64\rightarrow192$
& MLP; scale and shift outputs \\
Cross-attention
& $96$
& 1 layer; 32 heads; no dropout \\
Contact-occupancy head
& $160\rightarrow512\rightarrow512\rightarrow512\rightarrow1$
& MLP; sigmoid output \\
\hline
\multicolumn{3}{@{}l@{}}{\textit{MLP planner (833,322 parameters)}} \\
Shared backbone
& $939\rightarrow285\rightarrow285$
& MLP; ELU \\
Terrain-weight head
& $285\rightarrow1683$
& Reshape to $9\times187$; softmax \\
Contact-occupancy head
& $285\rightarrow9$
& Linear projection; sigmoid output \\
\end{tabularx}
\end{table*}

\subsubsection*{Observation encoding and curricula}

The policy height scans use a spacing of $0.1\,\mathrm{m}$,
with the spatial coverage listed in
Table~\ref{tab:supp_observation_curriculum}.
These policy observation grids are distinct from the
$0.03\,\mathrm{m}$ reconstruction grid used by the neural mapping module.

For nonzero goal commands, the horizontal component is
$\mathbf{d}_{xy}/\max(\|\mathbf{d}_{xy}\|,r)$,
where $\mathbf{d}_{xy}$ is the horizontal target displacement
in the body-yaw frame.
We use $r=0.3\,\mathrm{m}$ for reference-following and
$r=0.5\,\mathrm{m}$ for goal-reaching.
The heading component is a relative angle wrapped to $[-\pi,\pi)$.
During reference-following, the target is sampled uniformly
$0.5$--$1.5\,\mathrm{s}$ ahead, capped at the end of the reference clip.

In Stage~1, the initialization phase decreases by 0.05 after
an episode reaches phase 0.99 and increases by 0.05 otherwise,
with the result clipped to $[0,0.99]$.
Stage~2 uses uniformly sampled initialization phases.
For the Stage-3 goal-reaching curriculum, the per-environment
success estimate is updated after each episode as
$s\leftarrow0.8s+0.2I_{\mathrm{success}}$.
Promotion requires a successful episode and $s>0.5$.
An episode is successful for promotion if it reaches the goal
at the required height and completes without early termination.

\begin{table}[!htbp]
\centering
\caption{\textbf{Observation and curriculum settings.}
Reference phases are fractions of the motion-clip duration.
Horizontal curriculum distances are measured only when
the vertical goal error is below $0.1\,\mathrm{m}$.}
\label{tab:supp_observation_curriculum}
\begin{tabularx}{\columnwidth}{X|l}
Parameter & Value \\ \hline
\multicolumn{2}{l}{\textit{Observation encoding}} \\
Proprioceptive history & 5 control steps \\
Height-scan spacing & $0.1\,\mathrm{m}$ \\
Planner scan coverage & $1.6\times1.0\,\mathrm{m}$ \\
Tracker scan coverage & $0.8\times0.6\,\mathrm{m}$ \\
Reference-goal lookahead & $0.5$--$1.5\,\mathrm{s}$ \\
Goal scaling radius, reference-following & $0.3\,\mathrm{m}$ \\
Goal scaling radius, goal-reaching & $0.5\,\mathrm{m}$ \\
\hline
\multicolumn{2}{l}{\textit{Stage-1 initialization curriculum}} \\
Initial reference phase & $0.9$ \\
Completion threshold & $0.99$ \\
Phase adjustment & $0.05$ \\
Allowed initialization phase & $[0,0.99]$ \\
\hline
\multicolumn{2}{l}{\textit{Stage-3 terrain curriculum}} \\
Difficulty levels & 10 \\
Success-EMA update weight & $0.2$ \\
Success-EMA promotion threshold & $0.5$ \\
Horizontal promotion threshold & $0.5\,\mathrm{m}$ \\
Horizontal demotion threshold & $1.0\,\mathrm{m}$ \\
Vertical goal tolerance & $0.1\,\mathrm{m}$ \\
Maximum goal-reaching episode duration & $15\,\mathrm{s}$ \\
\end{tabularx}
\end{table}

\subsubsection*{Planner optimization and hindsight replay}

We optimize the planner loss in \eqref{eq:planner_loss}
with Adam, using unit coefficients for the affordance and
contact-occupancy losses.
Following planner initialization, each PPO iteration includes
ten planner optimizer updates.
Each update samples 4,096 replay entries and accumulates
gradients over eight sub-batches before taking one optimizer step.
Updates begin once at least 1,000 replay entries are available;
in distributed training, every worker must satisfy this condition.

When reference-following and goal-reaching replay buffers
are both nonempty, their contributions to a planner batch are
proportional to their stored sample counts.
Within each partitioned buffer, sampling is uniform over terrain
types, then over tiles of the selected type, and finally over
stored timesteps within the selected tile.

The trajectory recorder stores at most 500 control steps per episode.
For admitted goal-reaching episodes, the recorded segment is
further truncated at the first step within $0.2\,\mathrm{m}$
of the goal horizontally, when such a step exists.
This truncation removes the near-goal tail from planner supervision;
the $0.5\,\mathrm{m}$ horizontal and $0.1\,\mathrm{m}$ vertical
thresholds determine admission to the replay buffer.
For quality-based eviction, positive mechanical work is normalized
by the initial horizontal goal distance, with the denominator
clamped below at $0.5\,\mathrm{m}$.

\begin{table}[!htbp]
\centering
\caption{\textbf{Planner optimization and hindsight-labeling settings.}
Contact duration and trajectory length are measured in
$50\,\mathrm{Hz}$ control steps.
The trajectory-storage limit is separate from the
environment's episode timeout.}
\label{tab:supp_replay}
\begin{tabularx}{\columnwidth}{X|l}
Parameter & Value \\ \hline
\multicolumn{2}{l}{\textit{Optimization and decoding}} \\
Optimizer & Adam \\
Planner learning rate & $10^{-4}$ \\
Adam $(\beta_1,\beta_2)$ & $(0.9,0.999)$ \\
Adam $\epsilon$ & $10^{-8}$ \\
Weight decay & $0$ \\
Affordance / occupancy loss coefficients & $1.0$ / $1.0$ \\
Gaussian target bandwidth & $0.1\,\mathrm{m}$ \\
Planner updates per PPO iteration & 10 \\
Replay samples per optimizer update & 4,096 \\
Gradient-accumulation sub-batches & 8 \\
Minimum replay entries before updating & 1,000 \\
Contact-occupancy threshold & $0.5$ \\
Decoder $k$, quantitative evaluations & 1 \\
\hline
\multicolumn{2}{l}{\textit{Hindsight labeling and storage}} \\
Vertical contact-force threshold & $1\,\mathrm{N}$ \\
Minimum contact-segment duration & 10 steps ($0.2\,\mathrm{s}$) \\
Contact lookahead horizon & 75 steps ($1.5\,\mathrm{s}$) \\
Target footprint in body-yaw coordinates
& $[-0.8,0.8]\times[-0.5,0.5]\,\mathrm{m}$ \\
Maximum recorded steps per episode & 500 \\
Goal-reaching replay truncation distance & $0.2\,\mathrm{m}$ \\
Minimum distance used for quality normalization & $0.5\,\mathrm{m}$ \\
\end{tabularx}
\end{table}

\subsubsection*{Contact and reward annealing}
During Stage~2, contact-input selection and contact-reward weighting follow separate linear schedules:
\begin{equation}
\eta_{\xi}(u)=
\begin{cases}
0, & u \leq u_{\xi},\\
\dfrac{u-u_{\xi}}{N_{\mathrm{ann}}},
& u_{\xi}<u<u_{\xi}+N_{\mathrm{ann}},\\
1, & u \geq u_{\xi}+N_{\mathrm{ann}},
\end{cases}
\quad
\xi\in\{\mathrm{obs},\mathrm{rew}\}.
\label{eq:supp_contact_annealing}
\end{equation}
Here, $u$ is the PPO iteration counter, $u_{\mathrm{obs}}$ and
$u_{\mathrm{rew}}$ are the respective transition start iterations,
and $N_{\mathrm{ann}}$ is their shared transition duration.
The reward transition starts later, with
$u_{\mathrm{rew}}>u_{\mathrm{obs}}$.

The input schedule selects the complete reference or predicted contact command independently for each environment at each control step; contact coordinates are not interpolated. The reward schedule multiplies the three reference-contact terms in Table~\ref{tab:rew_func} by $1-\eta_{\mathrm{rew}}$ and the planner-contact term by $\eta_{\mathrm{rew}}$. Their resulting weights are $5.0(1-\eta_{\mathrm{rew}})$, $2.0(1-\eta_{\mathrm{rew}})$, $1.0(1-\eta_{\mathrm{rew}})$, and $1.5\eta_{\mathrm{rew}}$, respectively. All other Stage~2 reward weights remain unchanged. Stage~3 uses predicted contacts throughout, disables the reference-contact terms, and retains the planner-contact weight of $1.5$.

\subsubsection*{PPO parameters}

Table~\ref{tab:ppo} lists the PPO optimization settings.
Each iteration collects 24 control steps per environment
and performs five epochs with eight minibatches per epoch.
Advantages are normalized across the collected rollout;
with the Stage-3 dual critic, normalization is performed separately
for reference-following and goal-reaching samples.

The goal-reaching critic uses a fixed learning rate of $10^{-3}$.
During its first 500 PPO iterations, goal-reaching samples train
the new critic but are excluded from the actor's PPO surrogate loss.
Value-loss clipping is disabled for this critic during warmup.
The tracker and reference-following critic use the KL-adaptive
learning rate.

\begin{table*}[!htbp]
\centering
\caption{\textbf{PPO optimization settings.}
Batch sizes count transitions. Per-worker minibatches are one eighth of the corresponding rollout batch;
effective total minibatches include all workers. Stage-1 distributed totals are not reported.}
\label{tab:ppo}
\renewcommand{\arraystretch}{1.15}
\begin{tabularx}{\textwidth}{>{\raggedright\arraybackslash}p{0.52\textwidth}|*{3}{>{\centering\arraybackslash}X}}
Parameter & Stage~1 & Stage~2 & Stage~3 \\ \hline
\multicolumn{4}{l}{\textit{Rollout and batch sizes}} \\
Parallel environments per worker & $12{,}000$ & $2{,}400$ & $2{,}100$ \\
Rollout batch per worker & $288{,}000$ & $57{,}600$ & $50{,}400$ \\
Minibatch per worker & $36{,}000$ & $7{,}200$ & $6{,}300$ \\
Training workers (one GPU each) & --- & $2$ & $4$ \\
Total parallel environments & --- & $4{,}800$ & $8{,}400$ \\
Total rollout batch & --- & $115{,}200$ & $201{,}600$ \\
Effective total minibatch & --- & $14{,}400$ & $25{,}200$ \\
\hline
\multicolumn{4}{l}{\textit{Shared PPO settings}} \\
Rollout steps per environment & \multicolumn{3}{c}{$24$} \\
Optimization epochs & \multicolumn{3}{c}{$5$} \\
Minibatches per epoch & \multicolumn{3}{c}{$8$} \\
Policy clip range & \multicolumn{3}{c}{$0.2$} \\
Value-loss coefficient & \multicolumn{3}{c}{$1.0$} \\
Maximum gradient norm & \multicolumn{3}{c}{$1.0$} \\
Discount factor & \multicolumn{3}{c}{$0.99$} \\
GAE parameter & \multicolumn{3}{c}{$0.95$} \\
\hline
\multicolumn{4}{l}{\textit{Tracker and reference-following critic}} \\
Initial learning rate & \multicolumn{3}{c}{$10^{-4}$} \\
Learning-rate schedule & \multicolumn{3}{c}{KL-adaptive} \\
Target KL & $0.02$ & $0.01$ & $0.01$ \\
Entropy coefficient & 0.006 & $0.005$ & $0.002$ \\
\hline
\multicolumn{4}{l}{\textit{Stage-3 goal-reaching critic}} \\
Learning rate (fixed) & --- & --- & $10^{-3}$ \\
Warmup duration (PPO iterations) & --- & --- & $500$ \\
\end{tabularx}
\end{table*}

\subsection*{Evaluation Details}
\subsubsection*{Terrains}
\phantomsection
\label{sec:supp_evaluation_terrains}
\noindent Reference-anchored terrains: 
\begin{itemize}
    \item The same 123 reference configurations used for training
\end{itemize}

\noindent Goal-reaching terrains: 
\begin{itemize}
    \item The same 32 terrain configurations used in Stage~3 training, with difficulty level even sampled from the training range.
\end{itemize}

\begin{figure*}
\centering
\includegraphics[width=\linewidth]{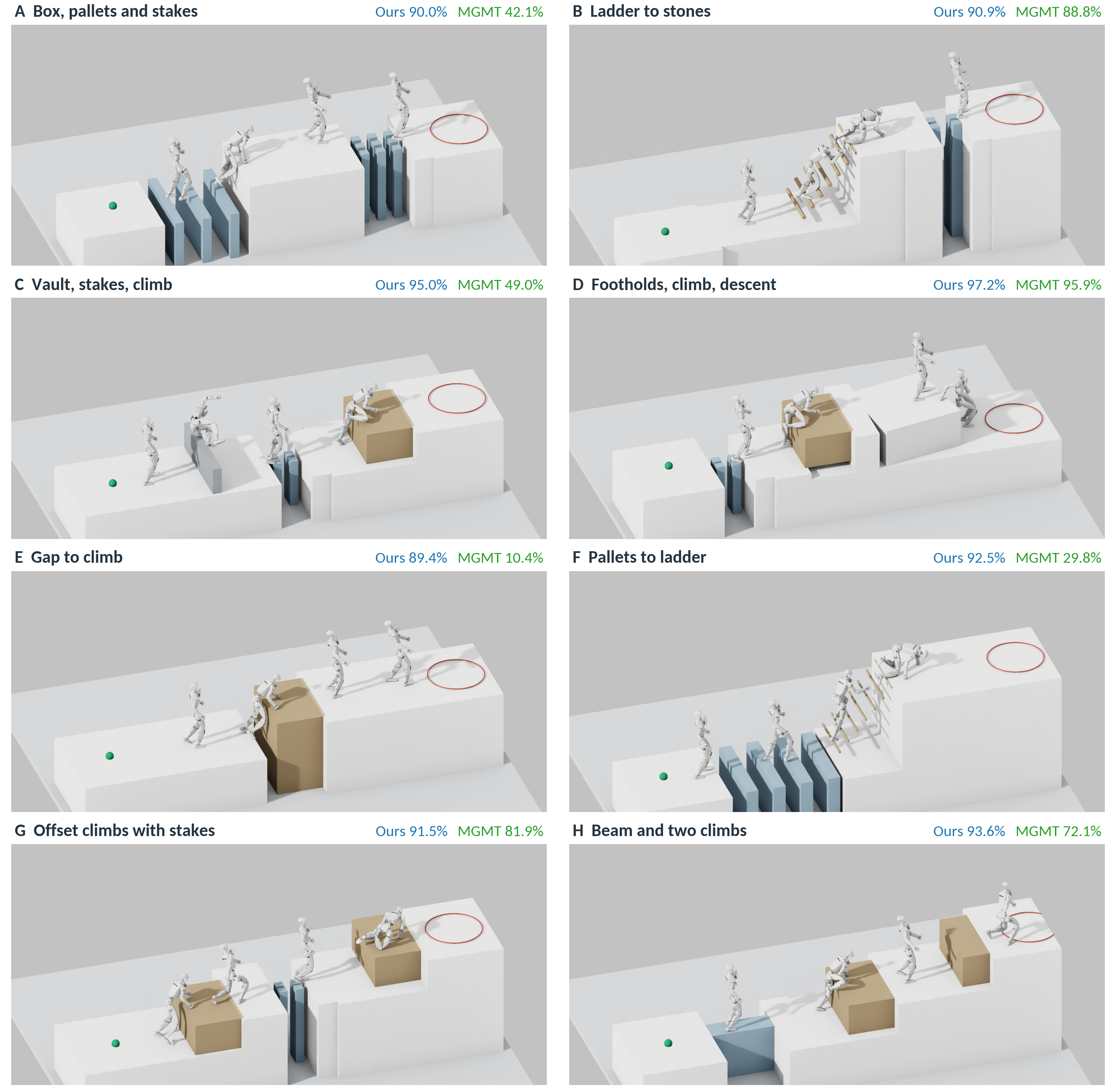}
\caption{Eight curated composite courses used for evaluation. Each panel shows four successive robot poses from one successful traversal. Green markers indicate starting positions, and red circles show the 0.5\,m horizontal goal tolerance. Success rates for Ours and MGMT are measured over 1,000 episodes per method and course.}
\label{fig:composite}
\end{figure*}

\noindent Unseen terrains
\begin{itemize}
\item \textbf{Compositions}
\begin{itemize}
\item Two randomly sampled goal-reaching terrains arranged sequentially.
\end{itemize}
\item \textbf{Distractors}
\begin{itemize}
\item Goal-reaching terrains with additional obstacles around the intended support area
\end{itemize}
\item \textbf{Roughness}
\begin{itemize}
\item Goal-reaching terrains with added irregular bumps on the surfaces, with maximum amplitude varying from 5 to 15 cm
\end{itemize}
\end{itemize}
\noindent \textbf{Curated composite courses:} A separate set of eight compact courses combines sparse footholds with whole-body obstacles and varied transition distances. These courses are shown in Figure~\ref{fig:composite} and evaluated separately from the group-level composition benchmark.



\end{document}